\documentclass[10pt,twocolumn,letterpaper]{article}

\usepackage[pagenumbers]{cvpr} 

\definecolor{cvprblue}{rgb}{0.21,0.49,0.74}
\usepackage[pagebackref,breaklinks,colorlinks,allcolors=cvprblue]{hyperref}
\usepackage{hyperref}

\def\confName{CVPR}
\def\confYear{2025}

\usepackage[utf8]{inputenc} 
\usepackage[T1]{fontenc}    
\usepackage{url}            
\usepackage{booktabs}       
\usepackage{amsfonts}      
\usepackage{nicefrac}       
\usepackage{microtype}      
\usepackage{xcolor}

\usepackage{graphicx}
\usepackage{booktabs}
\usepackage{algorithm}
\usepackage[noend]{algpseudocode}
\usepackage{caption}
\usepackage{subcaption}
\usepackage{xcolor}
\usepackage{multirow}
\usepackage{adjustbox}

\usepackage[sectionbib]{chapterbib}
\usepackage{wrapfig}

\usepackage{enumitem}

\newcommand{\bx}{\mathbf{x}}
\newcommand{\bw}{\mathbf{w}}
\newcommand{\bW}{\mathbf{W}}
\newcommand{\mbR}{\mathbb{R}}

\usepackage[accsupp]{axessibility}  

\usepackage{orcidlink}
\def\@onedot{\ifx\@let@token.\else.\null\fi\xspace}

\title{Self-Expansion of Pre-trained Models with Mixture of Adapters \\for Continual Learning}

\author{
Huiyi Wang$^{1,2}$, Haodong Lu$^{1}$, Lina Yao$^{2,1}$, Dong Gong$^1$\thanks{D. Gong is the corresponding author. This project was partially supported by an ARC DECRA Fellowship (DE230101591) to D. Gong, and PhD scholarship support from UNSW and CSIRO Data61. 
}\\
$^1$University of New South Wales, $^2$CSIRO's Data61\\
\texttt{\small \{huiyi.wang, haodong.lu, dong.gong\}@unsw.edu.au; lina.yao@data61.csiro.au}
}

\begin{document}
\maketitle

\begin{abstract}

Continual learning (CL) aims to continually accumulate knowledge from a non-stationary data stream without catastrophic forgetting of learned knowledge, requiring a balance between stability and adaptability. Relying on the generalizable representation in pre-trained models (PTMs), PTM-based CL methods perform effective continual adaptation on downstream tasks by adding learnable adapters or prompts upon the frozen PTMs. However, many existing PTM-based CL methods use restricted adaptation on a fixed set of these modules to avoid forgetting, suffering from limited CL ability. Periodically adding task-specific modules results in linear model growth rate and impaired knowledge reuse. We propose \textbf{S}elf-\textbf{E}xpansion of pre-trained models with \textbf{M}odularized \textbf{A}daptation (SEMA), a novel approach to enhance the control of stability-plasticity balance in PTM-based CL. SEMA automatically decides to reuse or add adapter modules on demand in CL, depending on whether significant distribution shift that cannot be handled is detected at different representation levels. 
We design modular adapter consisting of a functional adapter and a representation descriptor. The representation descriptors are trained as a distribution shift indicator and used to trigger self-expansion signals. For better composing the adapters, an expandable weighting router is learned jointly for mixture of adapter outputs. SEMA enables better knowledge reuse and sub-linear expansion rate. Extensive experiments demonstrate the effectiveness of the proposed self-expansion method, achieving state-of-the-art performance compared to PTM-based CL methods without memory rehearsal. Code is available at \url{https://github.com/huiyiwang01/SEMA-CL}.

\end{abstract}

\vspace{-0.2cm}

\section{Introduction}
\label{sec:intro}

With the development of deep neural networks, deep learning models have achieved significant success in various fields, such as computer vision \cite{resnet,vit}. 
However, real-world scenarios often present learning tasks in a dynamic data stream with non-stationary distributions \cite{mccloskey1989catastrophic}. Considering the need for efficient model updating and restricted budgets on storage and computation \cite{justus2018predicting}, it is not guaranteed to store all the historical data and repeatedly re-train the model. Continual learning (CL) is investigated to learn incrementally and accumulate knowledge efficiently from the non-stationary data stream without \emph{catastrophic forgetting}~\cite{lwf,nguyen2019toward} of previously learned knowledge \cite{icarl,progresscompress,de2021continual,wang2023comprehensive}. 
It requires CL approaches to achieve a balance between knowledge expansion (\ie, plasticity) and knowledge retention (\ie, stability) \cite{embracing, lmc, wang2023comprehensive}. 
Many CL approaches have been studied to tackle the challenge relying on different strategies, such as experience replay (ER) \cite{chaudhry2019tiny,agem,yan2022learning}, regularization on parameters or representations \cite{der_replay,yan2022learning,ewc}, and architectures with modularization or isolation \cite{yan2021dynamically,hat,lmc,mntdp,supsup}. 

\begin{figure*}[htp]
    \centering
    \includegraphics[width=0.9\linewidth]{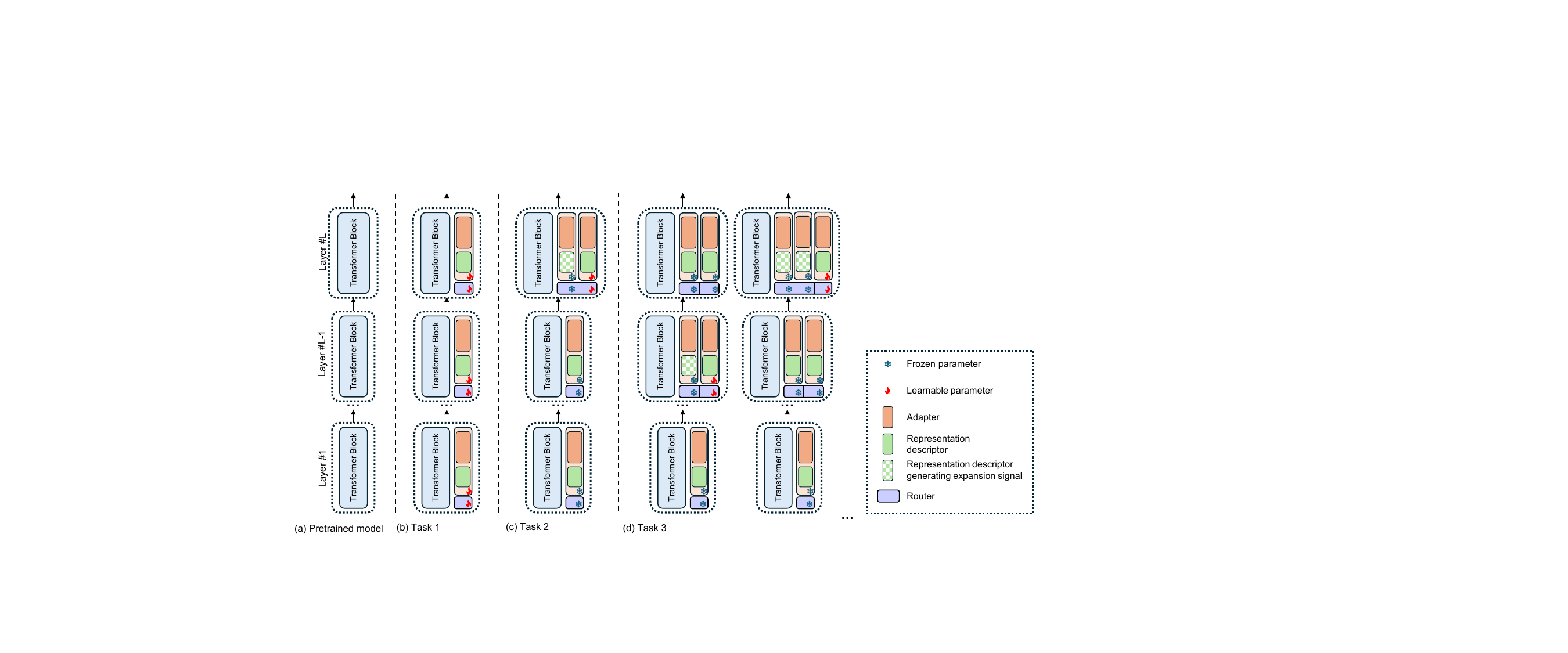}
    \vspace{-0.3cm}
    \caption{An example of the self-expansion process. (a) The PTM (\ie, ViT) with $L$ transformer layers at the initial point of CL. (b) The first session adaptation -- at Task 1, a modular adapter and a (dummy) router is added and trained in each transformer layer. 
    (c) The modular adapters and routers added in the previous step (Task 1) are frozen to alleviate forgetting. When Task 2 arrives, \emph{only} the representation descriptor in the $L$-th layer detects feature distribution shift (with novel patterns) and generates \emph{expansion signal}. A new module is added and trained in the $L$-th layer, with the router expanded and updated. (d) At Task 3, new adapter is added at $L-1$-th layer after the expansion signal is firstly generated. In this demo example, the expansion is triggered and produced again in the $L$-th layer, following the expansion in the $L-1$-th layer. If a task does not trigger expansion signal in any layer (implying no significantly different pattern), expansion would not happen, and existing adapters would be reused. More discussions are in Appendix \ref{supp:training_procedure}.
    }
    \label{fig:process}
    \vspace{-0.45cm}
\end{figure*}

\par
Given the progress in the pre-trained models (PTMs) with reliable representation, recent works explore the potential of using PTMs, such as Vision Transformer (ViT) \cite{vit}, as the starting point of CL, unlike the ``training-from-scratch'' paradigm. 
PTM-based CL approaches \cite{l2p,dualprompt} usually keep the PTMs frozen to enable stable representation and alleviate forgetting. The PTMs are continually adapted to downstream tasks through parameter-efficient fine-tuning with newly expanded parameters as prompts and/or adapters \cite{l2p,dualprompt,cmoa,coda,revisiting,proof,ranpac,boosting}. 
On the other hand, some methods enable continual fine-tuning of PTMs on real-world downstream tasks arriving in a streaming manner. 
Many PTM-based CL approaches mainly add and learn a \emph{fixed} set/pool of prompts \cite{coop,vpt} or adapters \cite{adaptformer} shared by all downstream tasks in the stream \cite{l2p,dualprompt,revisiting,ranpac}. 
To alleviate forgetting caused by the interference on the newly added parameters, they restrict the parameter updating only on the first task seen in stream \cite{revisiting,ranpac} or use various regularization on the shared parameters \cite{l2p,dualprompt}. Their continual adaptation potentials are limited by the fixed and static size of prompt and adapter parameters. Some recent methods expand the PTMs with task-specific parameters to produce input-conditioned prompts \cite{coda} or ensemble of adapters \cite{ease}. The task-specifically added modules can help reduce the interference but cause a \emph{linearly}-scaled model size (w.r.t. number of tasks) and restrained knowledge sharing and reuse.

Considering that the PTM and the newly added parameters in expansion can provide a \emph{stable} representation and a knowledge \emph{extension} mechanism for CL, respectively, we focus on how to further enhance the control of the \emph{stability-plasticity} balance during continual expansion. Although task-specific expansion of PTMs \cite{ease,coda} directly reduces the cross-task conflicts, it causes undesired \emph{linear} scaling of model size and may impair knowledge transfer/reuse \cite{progresscompress,lmc,mntdp}. To address these issues, we propose SEMA, a CL approach with \textbf{S}elf-\textbf{E}xpansion of pre-trained models with \textbf{M}odularized \textbf{A}daptation. It automatically expands PTMs with modularized adapters on demand and continually learns them to accommodate the distribution shifts without overwriting previously learned knowledge. Unlike existing methods that expand PTMs with a pre-defined fixed-size pool \cite{l2p,revisiting,ranpac,boosting} or task-specific components \cite{dualprompt,coda,ease}, we design modularized adapters to enable SEMA automatically decide \emph{when} and \emph{where} (\ie, which layer) to expand the PTM (\ie, a pre-trained ViT) on demand for tackling new requirements with sufficient and flexible plasticity, as shown in Fig. \ref{fig:process}. The model continually learns how to \emph{compose} the learned adapters. With the enhanced knowledge transfer and reuse, SEMA can thus perform better by only expanding the parameter size \emph{sub-linearly}. 

We introduce \emph{modular/modularized adapters} that can be identified and reused to solve new tasks, selectively adding and learning a subset of new adapters for unseen knowledge. Specifically, we design the modular adapter as a pair of a functional \emph{adapter} and a \emph{representation descriptor} (RD). The functional adapters produce specific feature representations to adapt to the different requirements of different tasks. The RDs are jointly trained to capture the \emph{feature distribution} relevant to the coupled adapter at corresponding layers, serving as indicators of distribution shift at the representation level of intermediate layers. SEMA can use the representation descriptors for self-expansion -- a new modular adapter is added at a specific layer if and only if all the representation descriptors indicate the input feature as a unseen pattern; otherwise, the existing frozen adapters are reused, resulting in \emph{sub-linear} expansion. They can be implemented as a model with density estimation or novelty detection ability, such as autoencoder (AE) \cite{ae} or variational autoencoder (VAE) \cite{vae}. The module expansion at each layer can happen flexibly to supplement existing representation space, leading to sufficient plasticity. The on-demand expansion strategy strengthens the knowledge transfer and reuse, compared to the task-specific expansion \cite{coda,ease}. For example, cat images and dog images have more shared features than food images; the SEMA model trained only on cat images tends to expand more new adapters when training on food images than on dog images. 
To effectively \emph{compose} the adapters, we design an \emph{expandable weighting router} to produce layer-wise weighted mixture of the adapters in a form of mixture of experts (MoE), which are expanded and learned in the self-expansion process. Despite the RDs may be used for adapter assignment by hard selection, the learned soft mixture router can perform more effectively (Appendix \ref{supp:router_rd}). We summarize our contributions as follows:
\begin{itemize}[itemsep=0cm,parsep=0.05cm]
    \item We propose a novel continual learning approach via self-expansion of PTMs with modularized adapters, \ie SEMA. In CL, it automatically determines the expansion necessity and location for new adapters, 
    adding them at specific layers to accommodate new patterns in samples.
    The model enhances the control of stability-plasticity trade-off through adapter reuse and flexible expansion performed only on demand. SEMA enables \emph{sub-linear} expansion and operates without the need for rehearsal.     
    \item To achieve SEMA, we introduce modular adapters comprising a functional adapter and a representation descriptor. The representation descriptor maintains the distribution of pertinent input features, serving as a local novel pattern detector for expansion during training. The expandable weighting router is maintained simultaneously for \emph{composing} the adapters via weighted mixture.
    \item Extensive experiments are conducted to validate the effectiveness and analyze the behavior of the proposed method, which demonstrates the model's ability on alleviating forgetting and knowledge transfer as well as the plausibility of the automated process. 
\end{itemize}

\section{Related Work}
\label{sec:related}
\vspace{-0.02cm}

\noindent \textbf{Continual Learning (CL).} 
The mainstream taxonomy classifies continual learning methods into three
categories: replay-based methods, regularization-based methods
and architecture-based methods~\cite{de2021continual,wang2023comprehensive}. Replay-based methods aim to alleviate catastrophic forgetting by retaining a memory buffer to store the information from old tasks for future replay~\cite{icarl,chaudhry2019tiny,der_replay,gem}. With simple intuition and effectiveness in preventing forgetting, these methods are limited by the size of the memory buffer and may also raise privacy concerns. An alternative approach is to implicitly maintain a generative model for producing pseudo-samples with similar distribution to old classes~\cite{shin2017continual,kemker2017fearnet,chenshen2018memory,rostami2019complementary,riemer2019scalable}. Regularization-based methods penalize significant changes to important parameters for seen tasks~\cite{ewc,zenke2017continual,nguyen2017variational,aljundi2018memory,ahn2019uncertainty,zeno2018task}, or consolidate the knowledge learnt from previous tasks with knowledge distillation~\cite{lwf,hou2018lifelong,lee2019overcoming,zhang2020class}. Instead of using all available parameters for all tasks, architecture-based methods allocate a subset of parameters dedicated to each task, which can be performed with task masking~\cite{hat,ke2021achieving,supsup,mallya2018piggyback} or dynamic architecture~\cite{yoon2017lifelong,li2019learn,hung2019compacting,yan2021dynamically,learn_to_grow,aljundi2017expert,mntdp,lmc,oddl,sedem}. These methods tend to achieve optimal performance with less forgetting as isolating parameters and growing capacity for novel tasks reduce task interference during training, however, they are mostly restricted to simple applications due to the complex model design. 

\noindent \textbf{Parameter-Efficient Fine-Tuning (PEFT).} Parameter-efficient fine-tuning methods train a small set of additional parameters rather than the entire pre-trained model, which reduces the demands placed upon computational resources. 
Prompt tuning modifies input tokens/prefixes via learnable prompts~\cite{li2021prefix,vpt}. LoRA~\cite{hu2021lora} injects low-rank matrices to approximate weight updates and avoids additional inference latency via re-parameterization, which has been further utilized as experts with mixture modeling in recent works~\cite{adamix,mole,loramoe,mocle}. Adapters introduced by \cite{houlsby2019parameter}, along with its variants~\cite{adaptformer,jie2022convpass}, insert lightweight learnable modules into the transformer. To enhance the efficacy of adapter learning, \cite{unipelt} investigates different insertion forms, and \cite{pfeiffer2020adapterfusion,ruckle2020adapterdrop,adaptersoup} explores the potential of adapter compositions.

\noindent \textbf{PTM-based CL.} 
Recent works adopt PTMs, such as ViT and CLIP, as the backbone in the CL
system to exploit its robust representational ability and enable further adaptation on downstream tasks~\cite{zscl,slca, convprompt, jha2024clap4clip}. PTM can serve as a feature extractor for prototypes, which can be used for classification with distance measurement~\cite{mi2022training,pelosin2022simpler,revisiting,ranpac}. PEFT techniques are also widely used to adapt PTMs in CL, including adaptation and prompting. 
L2P~\cite{l2p} and DualPrompt~\cite{dualprompt} apply a pool of prompts in CL through visual prompt tuning~\cite{vpt}. The prompt learning process is further improved by \cite{coda} with an attention mechanism and input-conditioned weights. ConvPrompt \cite{convprompt} adds parameter per task using linguistic knowledge from a large language model. Similar to prompt tuning in CL, some works also explore the use of a fixed set of adapters~\cite{ada,cmoa,moeadapter} or task-oriented expansion~\cite{ease,inflora} for better transfer of ViT to downstream CL tasks. \cite{lae} builds a unified framework incorporating both prompt and adapter-based methods. \cite{chen2023lifelong} adds experts in the pre-training of large language models (LLMs). 

\begin{figure*}[!t]
    \centering
    \includegraphics[width=\linewidth]{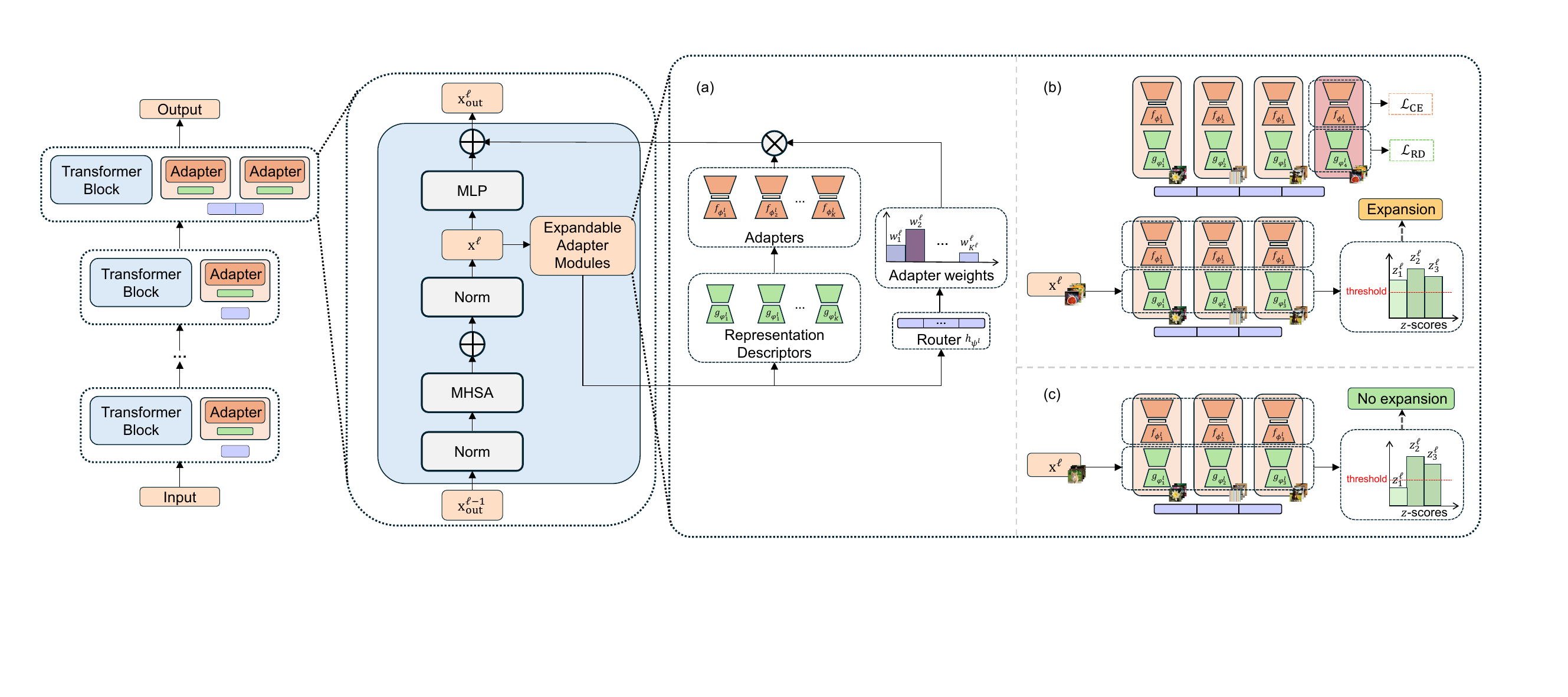}
    \vspace{-0.6cm}
    \caption{Overview of the model architecture. (a) shows the structure of expandable adapter modules with adapters, RDs and router. 
    (b) shows the scenario where expansion is triggered by representations with distribution different to previous tasks, estimated by RD. RDs are trained to align with the feature distribution of the corresponding task via only $\mathcal{L}_\text{RD}$, unaffected by gradients from the classification loss. (c) shows the scenario where incoming distribution can be handled by previously added modules, resulting in no expansion and adapter reuse.
    }
    \vspace{-0.45cm}
    \label{fig:sema_overview}
\end{figure*}

\section{Methodology}
\vspace{-0.02cm}
\label{sec:method}
\subsection{Problem Definition}
\vspace{-0.02cm}
Continual learning constructs a scenario where the model is required to learn from sequentially arriving tasks~\cite{de2021continual}. Consider a sequence of $T$ tasks $(\mathcal{D}^1, \mathcal{D}^2, ..., \mathcal{D}^T)$ with distribution shift, where $\mathcal{D}^t=\{(x_i^t, y_i^t)\}_{i=1}^{n_t}$ is the dataset containing $n_t$ data samples for the $t$-th task. Only the training samples from $\mathcal{D}^t$ are accessible while seeing the $t$-th task~\cite{l2p}, if without additional ER process \cite{chaudhry2019tiny}. In a typical class-incremental learning (CIL) scenario \cite{de2021continual}, the classes in different tasks are non-overlapping, specifically, with the label space of the $t$-th task denoted by $Y_t$, $Y_t \cap Y_{t'}=\emptyset$ for $t \neq t'$. Let $F_\theta: X\rightarrow Y$ (with $X$ and $Y$ denoting the domain of input and label) be a model parameterized with $\theta$. 
The goal of CL is to learn one model $F_\theta$ that can minimize the objective on each task $t$ in the stream: $\mathbb{E}_{(x,y)\in \mathit{D}^t} \mathcal{L}_\text{CE} (F_\theta(x), y)$, where $\mathcal{L}_\text{CE}(\cdot, \cdot)$ denotes the cross entropy loss in CIL.

\subsection{Overview}
We propose a PTM-based CL approach (\ie, SEMA) with a self-expansion mechanism to automatically add modularized adapters at arbitrary layers of the PTM (\ie, a pre-trained ViT with frozen parameters) on demand for handling automatically detected novel patterns in CL task stream, as shown in Fig. \ref{fig:process} and \ref{fig:sema_overview}. The proposed method simultaneously learns a weighted mixture router for composing the adapters for different inputs. The design enhances the balance of knowledge transfer/reuse and plasticity for handling novelty, with only \emph{sub-linear} expansion rate \cite{andreas2016neural,lmc}. 

\par
To achieve the modularized design of SEMA, we introduce the \emph{modular adapters} containing a pair of functional adapter $f_\phi(\cdot)$ and representation descriptor  $g_\varphi(\cdot)$, as defined in Sec. \ref{sec:adapter_formulation}. Each added functional adapter works as a branch of a specific layer of the pre-trained transformer; and the representation descriptor indicates the feature distribution that can be handled by the paired $f_\phi(\cdot)$. In CL, when new tasks arrive, $g_\varphi(\cdot)$'s of the already-added adapters are used to detect novel feature patterns layer-by-layer. Only when novel pattern (\ie, representation-level distribution shifts) are detected, new adapters, \ie, pairs of $(f_\phi(\cdot), g_\varphi(\cdot))$, are added and trained. After trained sufficiently, the adapters are kept frozen to alleviate forgetting and can be reused in future tasks. The details of the \emph{self-expansion strategy} are in Sec. \ref{sec:expansion_strategy}. At each layer of the PTM, an \emph{expandable weighting router} is continually maintained and updated for composing the adapters via weighted mixture, as introduced in Sec. \ref{sec:weighting}. When no adapters are added, the existing frozen adapters are retrieved and reused.

\subsection{Representation-Aware Modular Adapter}
\label{sec:adapter_formulation}
\noindent The modular adapter $(f_\phi(\cdot), g_\varphi(\cdot))$ is designed as a pair of \emph{functional adapter} $f_\phi(\cdot)$ and a \emph{representation descriptor} $g_\varphi(\cdot)$, which enables the module to be aware of the distribution of the local representation. One or more adapters can be added at arbitrary blocks/layers of the transformer.

\noindent \textbf{Functional adapter.} In a (pre-trained) ViT, there are $L$ layers of transformer blocks, where each of them mainly contains a multi-head self-attention (MHSA) module and a multi-layer perceptron (MLP) module \cite{vit}, as shown in Fig. \ref{fig:sema_overview}. We keep all the parameters in the ViT frozen and perform adaptation through the learnable parameters in the continually added adapters. As a commonly used solution \cite{adaptformer,revisiting}, the functional adapter with learnable parameters is added as a side branch of the MLP in any layer of the ViT. 
\par
Let $\bx^l\in\mbR^d$ denote the feature input of the MLP at $l$-th layer/block of ViT. In the proposed method, there can be different numbers (\ie, $K^l$) of adapters added at each layer through the self-expansion process. The $k$-th functional adapter at $l$-th layer is denoted as $f_{\phi^l_k}(\cdot)$. Each $f_{\phi^l_k}(\cdot)$ takes $\bx^l$ as input to bridge the representation gap between the pre-trained model and the downstream tasks. By default, we implement $f_{\phi^l_k}(\cdot)$ as a lightweight adapter \cite{adaptformer} containing a down-projection layer with parameters $\bW_{\text{down},k}^l\in \mathbb{R}^{d\times r}$, an up-projection layer with parameters $\bW_{\text{up},k}^l\in \mathbb{R}^{r\times d}$, and a non-linear ReLU activation~\cite{relu} in between.
By taking $\bx^l$ as input, the output of each functional adapter is formulated as 
\begin{equation} 
\label{eq:adaptformer}
	\textstyle
	f_{\phi^l_k}(\bx^l) = \text{ReLU}(\bx^l \cdot \bW_{\text{down},k}^l)\cdot \bW_{\text{up},k}^l,
\end{equation}
where $\phi^l_k\equiv\{\bW_{\text{up},k}^l, \bW_{\text{down},k}^l\}$ and $\bx^l$ is treated as row vector for notation simplicity. 
If there is only one adapter at the $l$-th layer (\ie, $K^l=1$), the output representation of the MLP is adjusted as $\bx^l_\text{out}=\text{MLP}(\bx^l)+f_{\phi^l_k}(\bx^l)$. 
SEMA can continually expand the model with more than one adapters if needed. The number of adapters at each layer is automatically determined on demand, with a rate that is sub-linear w.r.t. number of tasks. Although similar adapter formulation have been used to handle CL, they only perform adaptation on the first task using only one adapter \cite{revisiting,ranpac} or periodically expand the PTM using task-specific adapters \emph{linearly} \cite{ease}. In addition to Eq. \ref{eq:adaptformer}, the functional adapters can also be implemented as other forms, such as LoRA \cite{hu2021lora}, as discussed in Sec. \ref{sec:ablation}. 
\par
\noindent \textbf{Representation descriptor.}
The representation descriptor (RD) $g_{\varphi^l_k}(\cdot)$ is paired with the functional descriptor $f_{\phi^l_k}(\cdot)$ to capture the characteristics of the local representation. It is designed and trained to indicate what kind of input representation can be handled by the corresponding functional adapter at each specific layer. 
Representation descriptors can be implemented as any model with density estimation or novelty detection ability. For simplicity, we implement them as AE \cite{ae}, containing an encoder and a decoder. 
When a new pair of modular adapter is added at layer $l$, the RD $g_{\varphi^l_k}(\cdot)$ is trained by minimizing the reconstruction loss on all the features fed to $f_{\phi^l_k}(\cdot)$, \ie, $\mathcal{X}^l_k$:
\begin{equation}
\label{eq:AE_loss}
\mathcal{L}_{\text{RD},k}^l (x) = \sum\nolimits_{\bx\in \mathcal{X}_k^l } || \bx - g_{\varphi^l_k}(\bx) ||_2^2.
\end{equation}
In our expansion strategy (in Sec. \ref{sec:expansion_strategy}), when a new task $t$ arrives, at each $l$-th layer,  if all existing RDs detect significantly novel distributions (based on the $z$-score of reconstruction errors), the expansion signal is triggered. $f_{\phi^l_k}(\cdot)$ and $g_{\varphi^l_k}(\cdot)$ are trained on this task $t$ and then kept frozen in the future. $\mathcal{X}_k^l$ represents the input feature $\bx^l$ of all the samples in this new expansion-triggering task $t$. 

\subsection{Expandable Weighting Router for Mixture Usage of Adapters}
\label{sec:weighting}
By definition, the representation descriptor can be used to compose the adapters, as in similar modular networks. However, it heavily relies on the statistics of similar inputs in a batch \cite{lmc} and can be unreliable for individual inputs. We thus directly maintain and learn an \emph{expandable weighting router} for a weighted mixture of the functional adapters. 

\par
For any $l$-th layer with $K^l$ adapters, the routing function is defined as $h_{\psi^l}(\cdot):\mbR^d\rightarrow \mbR^{K^l}$. Similar to \cite{loramoe}, we implement $h_{\psi^l}(\cdot)$ as a linear mapping function followed by a softmax operation $\bw^l = h_{\psi^l}(\bx^l) \equiv \text{softmax}( \bx^l \cdot \bW_{\text{mix}}^l )$, where $\bW_{\text{mix}}^l\in \mbR^{d\times K^l}$ is the parameter of $\psi^l$. 
As shown in Fig. \ref{fig:sema_overview}, the weights $\bw^l\in \mbR^{K^l}$ can produce the mixture of the added functional adapters to produce the output representation of the MLP in the transformer:
\begin{equation}
    \bx^l_\text{out}=\text{MLP}(\bx^l)+\sum\nolimits_{k=1}^{K^l} w_k^l\cdot f_{\phi^l_k}(\bx^l).
\end{equation}
When new adapter is added at any layer $l$, the router $h_{\psi^l}(\cdot)$, \ie, $\bW_{\text{mix}}^l$, is expanded for producing weights with one more dimension. The expanded router is trained together with the added adapters. While expanding the router, the parameters corresponding to the existing adapters remain frozen and only the newly added ones (\ie, a newly added column in $\bW_{\text{mix}}^l$) are trained. This approach, similar to the common practice for training classification heads in CL \cite{coda,inflora}, controls and restricts forgetting in the expandable router (shown in Fig. \ref{fig:adapter_routing}), though it cannot fully eliminate it.

\begin{table*}[t]
	\centering
      \resizebox{0.85\linewidth}{!}{
		\begin{tabular}{@{}lccccccccc cccccccc}
			\toprule
			\multicolumn{1}{l}{{\small Method}} & 
			\multicolumn{2}{c}{\small CIFAR-100} 
            & \multicolumn{2}{c}{\small 5-Task IN-R}
            & \multicolumn{2}{c}{\small 10-Task IN-R}
            & \multicolumn{2}{c}{\small 20-Task IN-R} 
            & \multicolumn{2}{c}{\small ImageNet-A}
			 & \multicolumn{2}{c}{\small VTAB} \\
		 & ${\bar{\mathcal{A}}}$ & {$\mathcal{A}_N $}
         & ${\bar{\mathcal{A}}}$ & {$\mathcal{A}_N $}
         & ${\bar{\mathcal{A}}}$ & {$\mathcal{A}_N $}
         & ${\bar{\mathcal{A}}}$ & {$\mathcal{A}_N $}
       & ${\bar{\mathcal{A}}}$  & {$\mathcal{A}_N $} 
       & ${\bar{\mathcal{A}}}$   & {$\mathcal{A}_N $} 
		\\
		\midrule
FT Adapter         & 47.88          & 30.9           & 53.91             & 41.23          & 45.31              & 30.93             & 38.51              & 24.22             & 29.78             & 17.64          & 59.98          & 43.50          \\
L2P                      & 84.77          & 77.87          & 77.40             & 73.59          & 66.97              & 62.72             & 70.67              & 62.90             & 47.16             & 38.48          & 81.19          & 80.83          \\
DualPrompt               & 86.60          & 80.43          & 76.39             & 72.29          & 72.83              & 66.75             & 62.33              & 61.97             & 59.54             & 50.23          & 82.89          & 79.79          \\
CODA-P                   & \textbf{91.55} & 86.11          &     81.63              & 76.98          & 81.11              & 75.25             & 75.00              & 70.02             & 47.29             & 35.02          & 79.88          & 81.58          \\
SimpleCIL                & 82.31          & 76.21          & 65.83             & 61.31          & 67.09              & 61.35             & 67.59              & 61.35             & 60.05             & 49.24          & 85.29          & 83.61          \\
ADAM                     & 90.55          & 85.62          & 79.91             & 74.25          & 79.11              & 73.15             & 75.84              & 69.10             & 60.15             & 49.24          & 85.29          & 83.61          \\
InfLoRA                & 90.51          & 85.05          & 78.58             & 72.58          & 81.39              & 75.32             & 78.87              & 72.60              & 59.71             & 46.21          & 88.90           & 87.63          \\
\midrule
SEMA                     & 91.37 & \textbf{86.98} & \textbf{84.75}    & \textbf{79.78} & \textbf{83.56}     & \textbf{78.00}    & \textbf{81.75}     & \textbf{74.53}    & \textbf{64.53}    & \textbf{53.32} & \textbf{91.26} & \textbf{89.64} \\
\bottomrule
    \end{tabular}}
    \vspace{-0.15cm}
    \caption{Comparison with ViT-based CL methods in CIL.  All models adopt ViT-B/16-IN1K as the backbone.}
     \label{tab:cil_result}	
\end{table*}

\begin{figure*}[ht]
\vspace{-0.2cm}
  \centering
  \begin{subfigure}[b]{0.22\linewidth}
    \includegraphics[width=\linewidth]{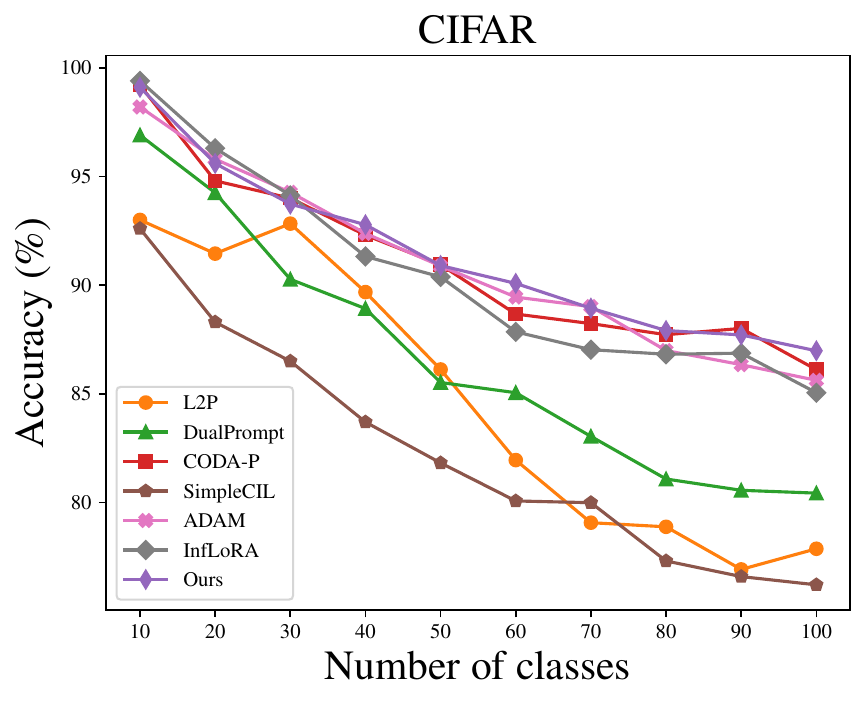}
  \end{subfigure}
  \begin{subfigure}[b]{0.22\linewidth}
    \includegraphics[width=\linewidth]{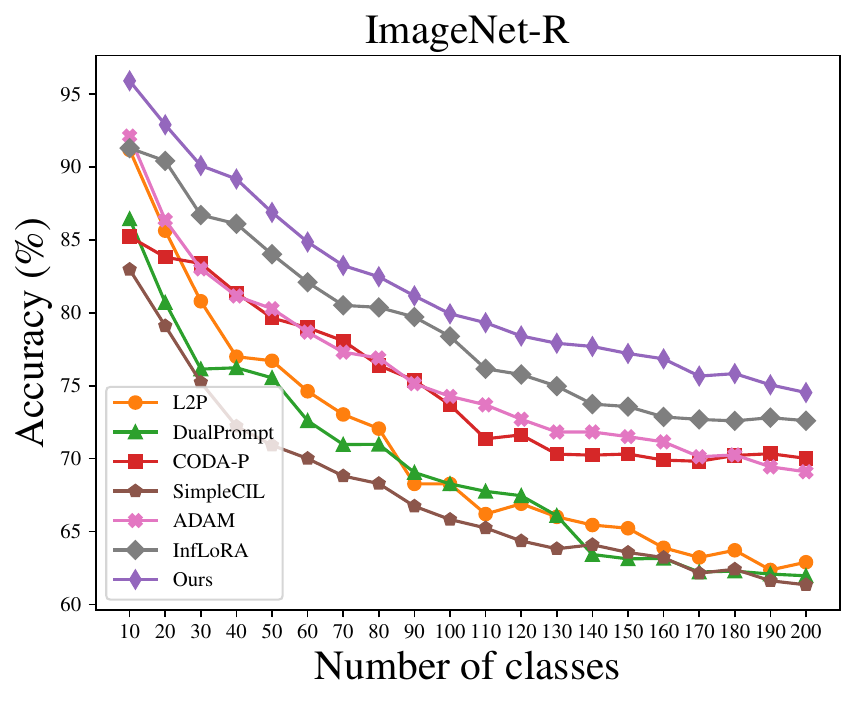}
  \end{subfigure}
  \begin{subfigure}[b]{0.22\linewidth}
    \includegraphics[width=\linewidth]{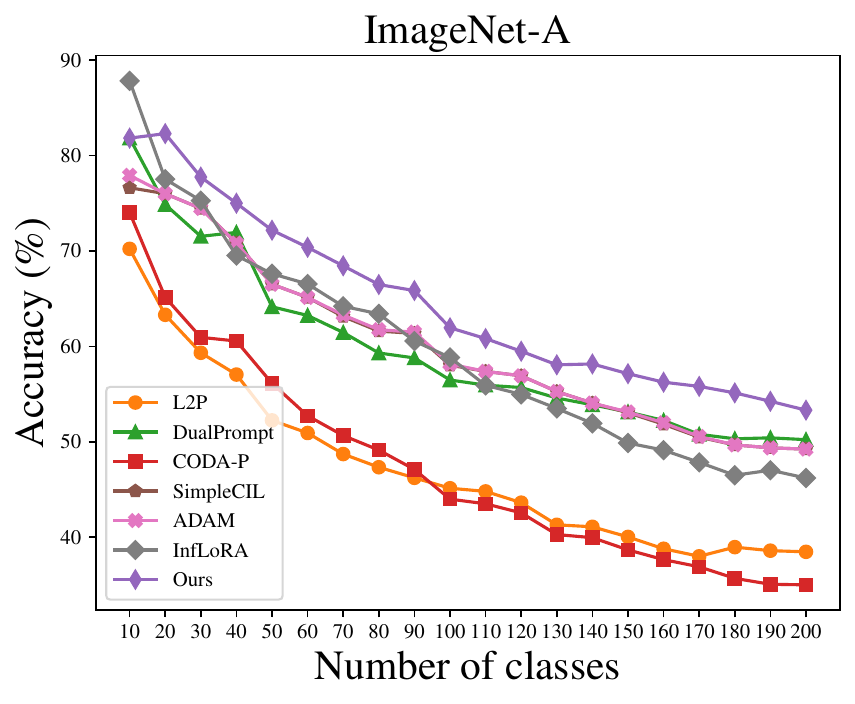}
  \end{subfigure}
  \begin{subfigure}[b]{0.22\linewidth}
    \includegraphics[width=\linewidth]{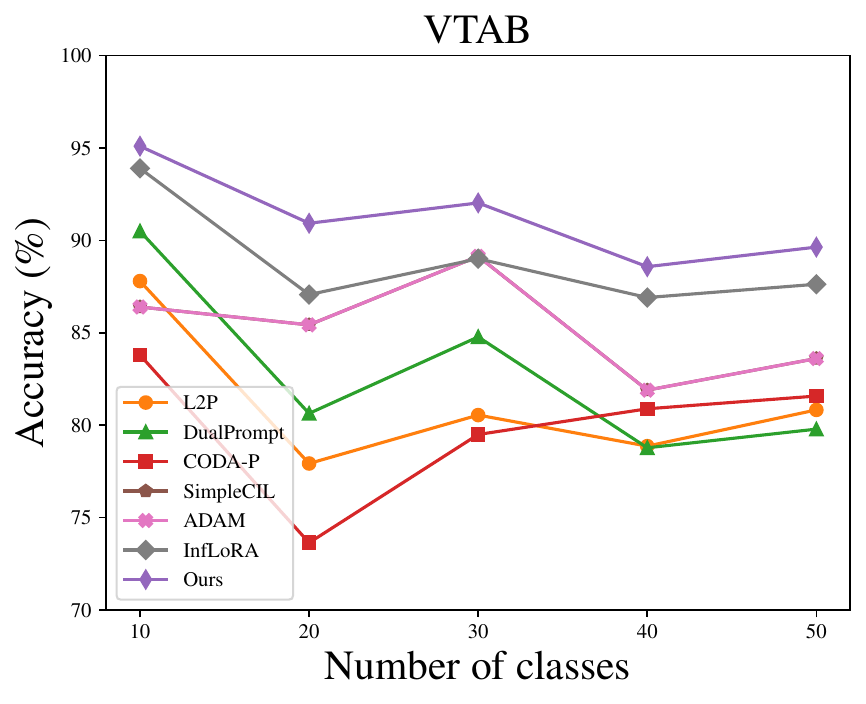}
  \end{subfigure}
  \vspace{-0.2cm}
  \caption{Incremental performance of different methods on class-incremental learning benchmarks. 
  }
  \vspace{-0.5cm}
  \label{fig:acc_record}
\end{figure*}

\subsection{Continual Learning Objective of SEMA}
In SEMA, the model $F_\theta(\cdot)$ for solving the tasks consists of learnable parameters from the functional adapters and router with learnable parameters, \ie, $\{\phi_k^l\}$ and $\{\psi^l\}$. The learnable parameters are dynamically added and learned. The representation descriptors are learned jointly for maintaining a state of the local representation. The overall objective in SEMA optimizes all these parameters:
\begin{equation}
\begin{aligned}
\min_{ \{\phi_k^l\}, \{\psi^l\}, \{\varphi^l_k\} } &\sum\nolimits_{t=1}^T \mathbb{E}_{(x,y)\in  \mathit{D}^t} \left[ \mathcal{L}_\text{CE} (F_{\{\phi_k^l\}, \{\psi^l\} } (x),~ y) \right. \\
+ & \left. \sum\nolimits_{l=1}^L \sum\nolimits_{k=1}^{K^l} \mathcal{L}_{\text{RD},k}^l (x; \varphi^l_k) \right].
\end{aligned}
\end{equation}

Learning of modular adapters is executed only when new modules are added. The learned modules are kept frozen to prevent forgetting. Optimization of RDs can be parallel to other parameters. If no module is added in a specific task due to no significant pattern being identified by RDs, the existing modules can be reused without training.

\subsection{Self-Expansion Strategy}
\label{sec:expansion_strategy}
The RDs provide the capacity to decide when and where to expand the model. We designed a more specific strategy to achieve the reliable self-expansion in the CL task stream.
\par
\noindent\textbf{Task-oriented expansion.} The expansion may occur at any time as new samples are seen during training. To incorporate the task identification prior knowledge in CL, especially CIL, we improve parameter efficiency and expansion stability with task-oriented expansion. We restrict the addition to at most one adapter per layer for each task. When a new task $t$ arrives, the method scans all samples in the \emph{first epoch} to decide whether to expand the model. If the expansion signal is triggered, only one adapter is added and then trained for the whole task; otherwise, the task $t$ data can reuse learned modules and the learning process moves to the next task.
\par
\noindent\textbf{$z$-score based expansion signal.}
When scanning through the new task data, an expansion signal at layer $l$ is triggered when significantly new patterns are identified. It reflects that a $\bx^l$ is out of the scope of all RDs, \ie, reconstruction error is high with each $g_{\varphi^l_k}(\bx)$ \cite{gong2019memorizing}, as illustrated in Fig. \ref{fig:expansion_signal}. However, it is impractical to directly use reconstruction error due to the perturbation and heterogeneous characteristics of each task and adapter. We thus compute and maintain the running statistics $\mu_k^l$ and standard deviation $\sigma_k^l$ of reconstruction error on all relevant inputs used in training. 
Given any $x^l$ in the scanning process for a future task, the $z$-score corresponding to each existing RD can be calculated as $z_k^l={(r_k^l-\mu_k^l)}/{\sigma_k^l}$ with $r_k^l$ as reconstruction error. If all $z_k^l$'s for $k=1,...,K^l$ are larger than a threshold, the expansion signal is triggered. Considering that the $z$-score has normalized out  perturbation and scale, the process can be very robust to the threshold setting, as shown in Sec. \ref{sec:ablation}.

\par
\noindent\textbf{Multi-layer expansion.}
We facilitate self-expansion across multiple layers through distinct decision processes. Upon encountering a new task, self-expansion operations are executed sequentially from shallow layers to deeper layers. As new adapters are introduced at shallow levels, training ensures representations are aligned accordingly. Subsequently, the model determines whether to continue expanding into subsequent layers. The adaptable multi-layer expansion facilitates the accommodation of various distribution shifts and enables flexible inter-class knowledge sharing \cite{surgical_finetuning, gao2024higher}.

\section{Experiments}
\label{sec:experiments}

\subsection{Setting and Implementation Details}
\label{sec:exp}
\textbf{Datasets.} Experiments are conducted on common datasets used for pre-trained ViT-based CIL: CIFAR-100~\cite{cifar_dataset}, ImageNet-R (IN-R)~\cite{inr_dataset}, ImageNet-A~\cite{ina_dataset} and VTAB~\cite{vtab_dataset}. 

\noindent \textbf{Baselines.} We validate our method by comparing with PTM-based rehearsal-free CL approaches using similar backbone (\eg, ViT) and methodology, including fully fine-tuning of the adapter, L2P~\cite{l2p}, DualPrompt~\cite{dualprompt}, CODA-P~\cite{coda}, SimpleCIL~\cite{revisiting}, ADAM with Adapter~\cite{revisiting} and InfLoRA~\cite{inflora}. 

\noindent \textbf{Training details.} We use the commonly used ViT-B/16 model \cite{vit} weights pre-trained on ImageNet-1K \cite{imagenet} as the PTM weights. We also conducted experiments with other pre-trained weights and left discussions in Appendix \ref{supp:21k_weight}. The batch size is set to 32. SGD is used as the optimizer with the initial learning rate set to 0.005 and 0.01 for adapters and RDs, respectively, decaying with cosine annealing. The hidden dimension of adapter is 16. In experiments, by default, we enable self-expansion in the last three transformer layers for simplicity 
without losing generality.

\subsection{Experimental Results}
We validate the proposed method by comparing with previous related state-of-the-art methods and reporting the average accuracy of all tasks $\mathcal{A}_N$~\cite{agem} and average incremental accuracy $\bar{\mathcal{A}}$~\cite{icarl} metrics in Tab. \ref{tab:cil_result}. 
It shows that our method performs better than other related methods in terms of the average accuracy at the last step $\mathcal{A}_N$, which reflects the final goal of CL. Fig. \ref{fig:acc_record} shows the variation in accuracy during the continual learning process.It shows the consistently superior performance of SEMA in the process. 
Although most previous approaches exhibit strong performance on CIFAR-100, the proposed methods shows more improvements on datasets containing adversarial samples similar to those found in ImageNet, due to its better stability-plasticity balance. 

\subsection{Ablation Studies and Analyses}
\label{sec:ablation}
\noindent\textbf{Ablation studies on module expansion and adapter composing.} 
We conduct ablation studies to demonstrate the effectiveness of the self-expansion process and investigate the influence of different adapter composing strategies, with the results reported in Tab. \ref{tab:ablation_expansion_selection}.  
We first conduct an experiment by removing the self-expansion process and only keeping the first-session adaptation (No Exp.), which is similar to ADAM \cite{revisiting} with slight difference on implementation. The results show that the self-expansion can work reliably to continually improve the adaptation results. 
\begin{table}[!h]
\vspace{-0.3cm}
	\centering
      \resizebox{0.7\linewidth}{!}{
		\begin{tabular}{@{}lccccccccc cccccccc}
			\toprule
			{{Method}} & 
   \multicolumn{2}{c}{ImageNet-A} & \multicolumn{2}{c}{VTAB} \\
        & ${\bar{\mathcal{A}}}$ & {$\mathcal{A}_N $} 
        & ${\bar{\mathcal{A}}}$  & {$\mathcal{A}_N $} 
		\\
		\midrule
  	SEMA & \textbf{64.53}       & \textbf{53.32}       & \textbf{91.26}       & \textbf{89.64} \\
        \midrule
        No Exp.  & 61.20 & 49.90 & 86.21 & 83.66 \\ \midrule 
        Avg. W. & 56.88 & 44.31 & 90.84 & 89.14 \\
        Rand. W. & 62.95 & 49.77 & 88.87 & 85.17 \\

        Top-1 Sel. & 62.00 & 50.56 & 90.83 & 88.61 \\
        Rand. Sel. & 61.70 & 50.36 & 90.82 & 88.51 \\

        \midrule
        Top-1 Sel. Inf.  &  61.96 & 50.36 & 90.95 & 88.84 \\
	\bottomrule
    \end{tabular}
    }
    \vspace{-0.2cm}
    \caption{Ablation studies on adapter expansion and composing.}
     \label{tab:ablation_expansion_selection}
     \vspace{-0.3cm}
\end{table}

\par
To demonstrate the benefits of the weighted mixture routing, we investigate several variants of SEMA with different adapter composing strategies. Firstly, we study two variants with a soft mixture of adapters relying average weighting (Avg. W.) and random weighting (Rand. W.), respectively. Tab. \ref{tab:ablation_expansion_selection} shows that the expandable weighting router learns an effective weighting function. 
We further study the variants that perform routing by selecting only a single adapter indicated by the highest value from the learned weighting router (Top-1 Sel.) or through random drawing (Rand. Sel.). Additionally, we evaluate SEMA trained with mixture routing, using an inference strategy that selects only the adapter with the highest weight (Top-1 Sel. Inf.).
The results show that the weighted soft mixture of the learned adapters works more effectively by encouraging the better usage of the learned adapters. More experiments about adapter composing using representation descriptor are in Appendix \ref{supp:router_rd}.

\begin{figure}
    \centering
    \includegraphics[width=0.7\linewidth]{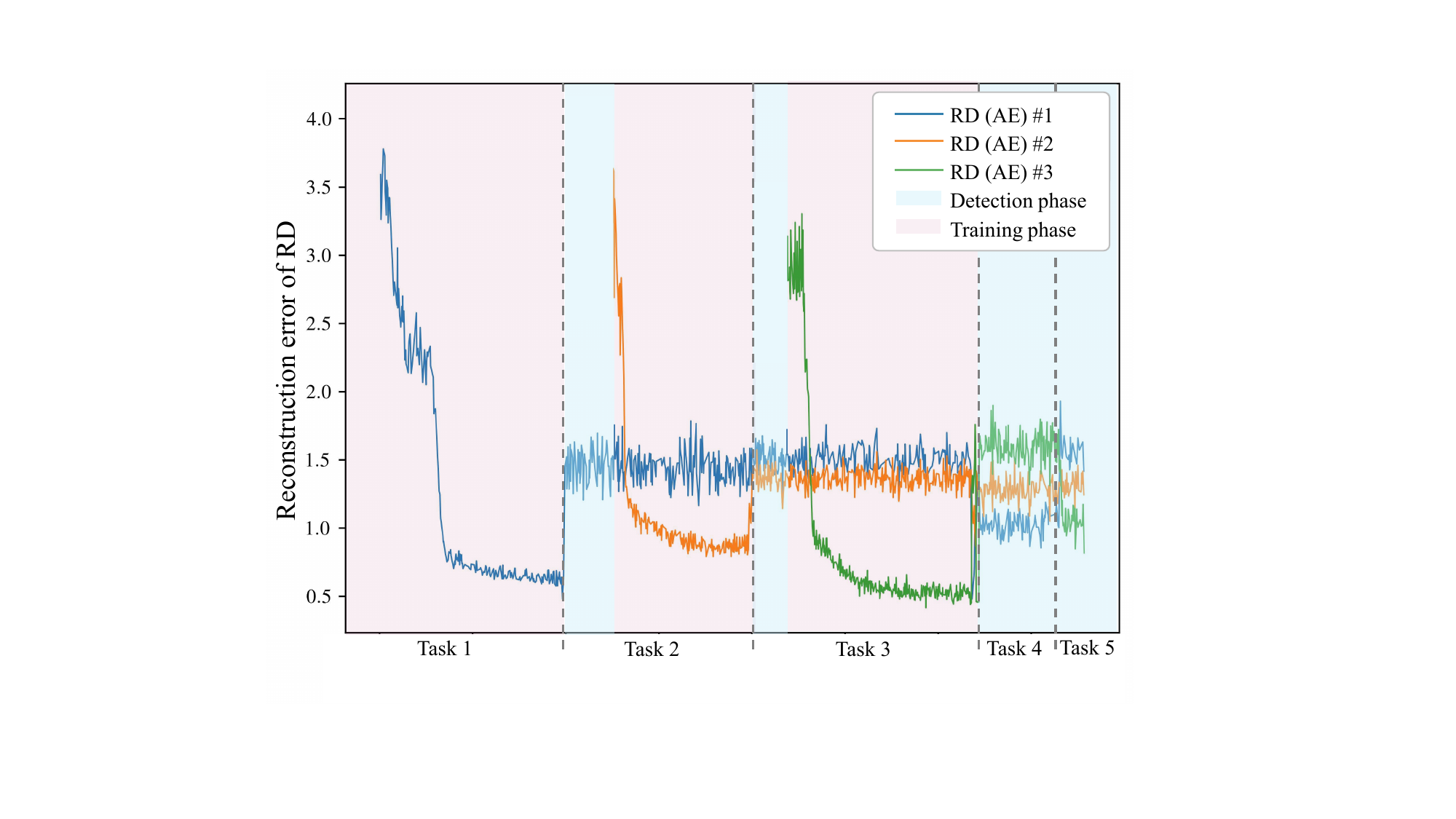}
    \vspace{-0.2cm}
    \caption{Reconstruction error during training to show the dynamic expansion process. Expansion occurs for Tasks 1, 2, and 3, while no expansion is triggered for Tasks 4 and 5 due to no detected distribution shift.}
    \label{fig:expansion_signal}
    \vspace{-0.3cm}
\end{figure}

\noindent\textbf{Analysis on dynamic expansion process.} 
To demonstrate how the representation descriptors are learned and how they work for self-expansion in CL, we visualize the reconstruction error of each AE-based RD corresponding to each sample seen during training, \ie, their representation features at specific layer, in Fig. \ref{fig:expansion_signal}. For more intuitive visualization and simplified experiment, in this analysis, we restrict the automatic self-expansion only to the last layer of transformer. The analysis is conducted on VTAB dataset. 
In this case shown in Fig. \ref{fig:expansion_signal}, the reconstruction error of each RD decreases and converges after training on the corresponding task, after the RD is added for handling this task. When a new task arrives, the reconstruction errors for the existing RDs are calculated and used to detect novelty. The expansion signal is generated when significantly high reconstruction errors (scaled as $z$-scores) are detected from all the previous RDs (in Task 2 and 3). In Task 4 and 5, all samples can be well covered by at least one previous RD, which implies no significant distribution shift is detected and results in no expansion. Note that the $z$-score (\ie, a normalized version of reconstruction error) is used for expansion in SEMA.

\begin{figure}[htp]
    \centering
    \includegraphics[width=0.65\linewidth]{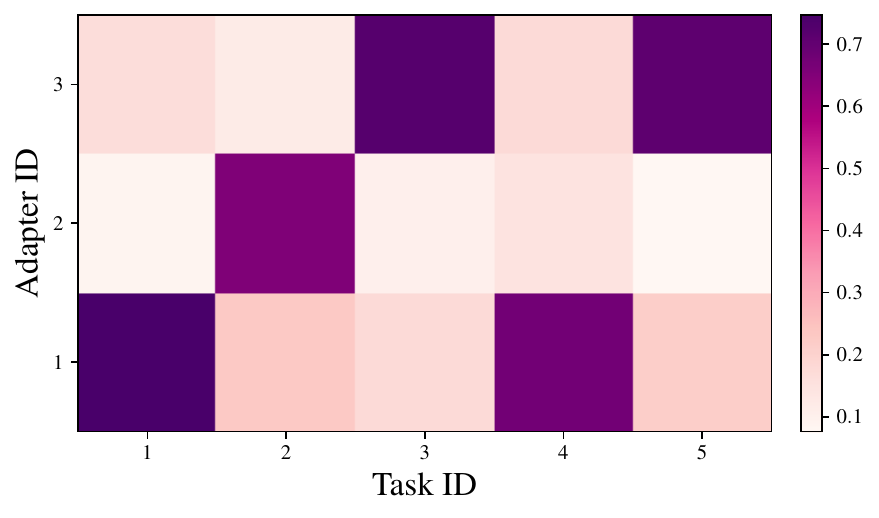}
    \vspace{-0.33cm}
    \caption{Visualization of adapter usage on VTAB. Adapters 1, 2, and 3 are added and trained on Tasks 1, 2, and 3, respectively. Tasks 4 and 5 primarily reuse Adapters 1 and 3 due to similar feature distributions with Tasks 1 and 3.}
    \label{fig:adapter_routing}
    \vspace{-0.5cm}
\end{figure}

\noindent\textbf{Analysis on adapter usage.}
Fig. \ref{fig:adapter_routing} demonstrates the average adapter usage of each task from VTAB. This analysis 
is produced by restricting self-expansion to the last layer, as in Fig. \ref{fig:expansion_signal}. Self-expansion is automatically produced for Task 1, 2 and 3. For tasks that triggered expansion, the adapters used are primarily those they were trained with, as shown in the figure. Task 4 and 5 share a similar selection pattern with the tasks they are similar with (Task 1 and 3 respectively), showing that added adapters are effectively reused for new tasks. More details are in Appendix \ref{supp:router_rd}.

\begin{figure}[!t]
  \centering
  \begin{subfigure}[b]{0.45\linewidth}
    \includegraphics[width=\linewidth]{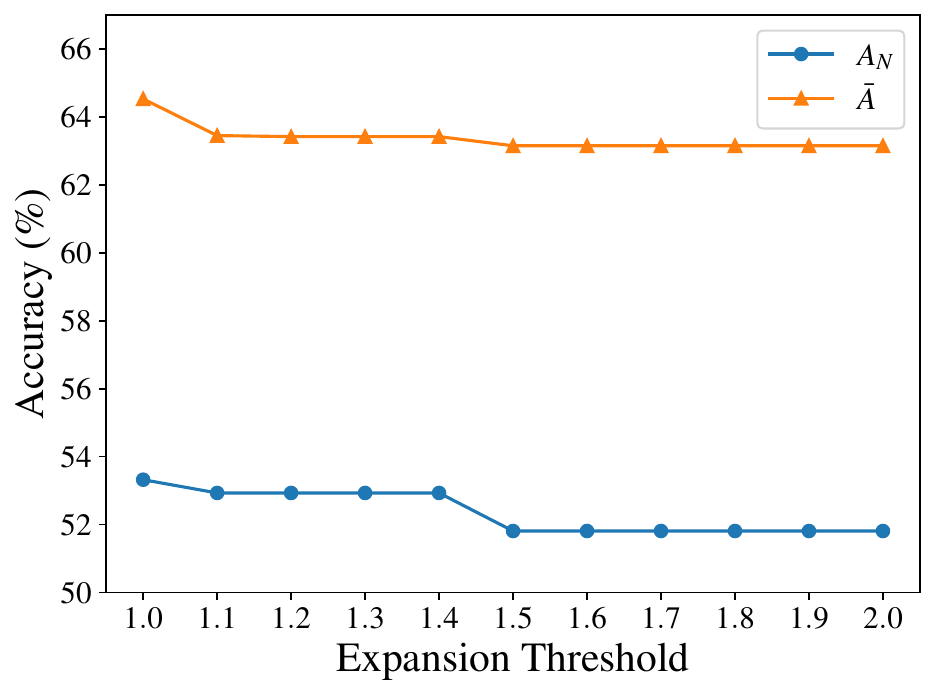}
    \caption{Accuracy }
    \label{fig:ina_acc_dev}
  \end{subfigure}
  \hfill
  \begin{subfigure}[b]{0.45\linewidth}
    \includegraphics[width=\linewidth]{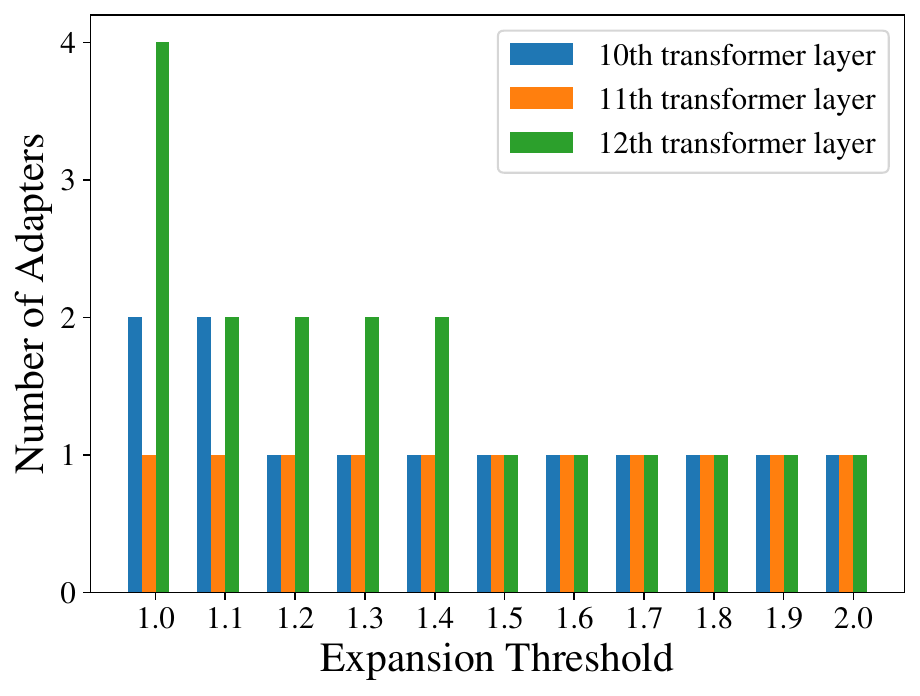}
    \caption{Num. of adapters}
    \label{fig:ina_num_adapter_dev}
  \end{subfigure}
  \vskip0.5\baselineskip
  \begin{subfigure}[b]{0.45\linewidth}
    \includegraphics[width=\linewidth]{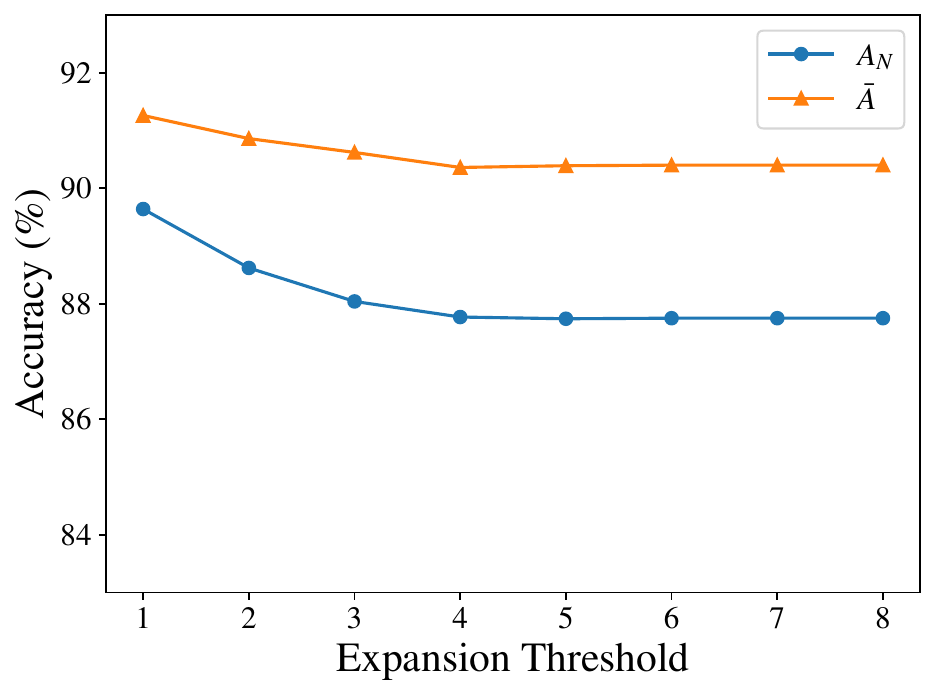}
    \caption{Accuracy }
    \label{fig:vtab_acc_dev}
  \end{subfigure}
  \hfill
  \begin{subfigure}[b]{0.45\linewidth}
    \includegraphics[width=\linewidth]{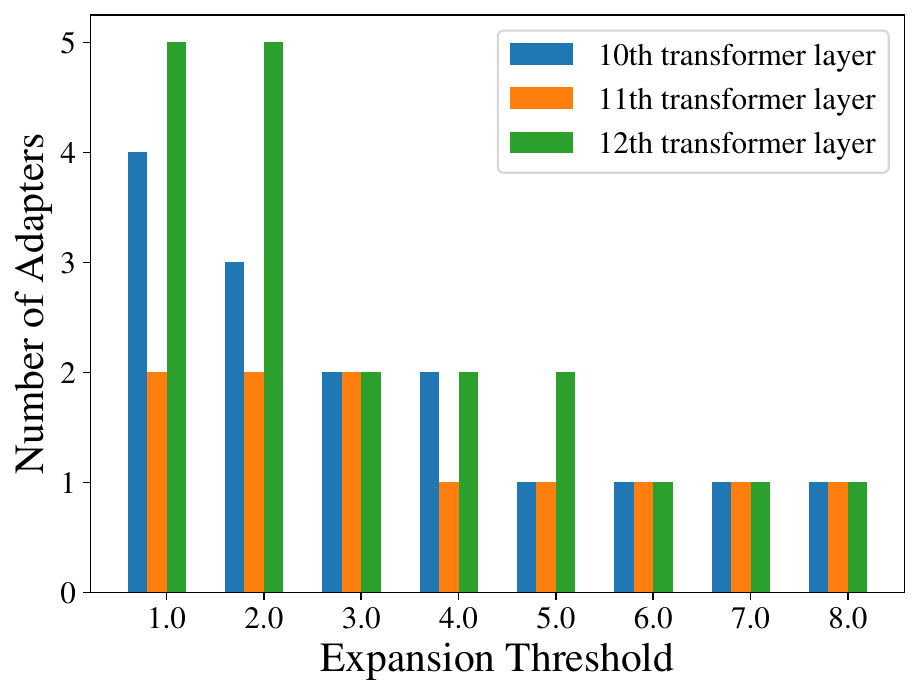}
    \caption{Num. of adapters}
    \label{fig:vtab_num_adapter_dev}
  \end{subfigure}
  \vspace{-0.15cm}
  \caption{
  Analysis of the impact of expansion threshold with (a)(b) ImageNet-A and (c)(d) VTAB. (a) and (c) show that SEMA can produce good accuracy stably with slight variation \wrt varying expansion threshold. (b) and (d) report how the number of added adapters (on the specific Transformer layers $\#10$, $\#11$, $\#12$) changes with the varying threshold values, corresponding to (a) and (c), respectively. 
  The proposed method is insensitive to the threshold. Adding more adapters may lead to higher accuracy, a proper threshold can achieve a balance between performance and model size.}
  \label{fig:ablate_expansion_threshold}
\end{figure}

\noindent\textbf{Study of expansion threshold.}
We investigate the impact of the expansion threshold on accuracy and the number of added adapters using ImageNet-A and VTAB. Firstly, the results 
in Fig. \ref{fig:ablate_expansion_threshold} show that the proposed method is not sensitive to the setting of the threshold, benefiting from the $z$-score-based expansion signal. Fig. \ref{fig:ina_num_adapter_dev} and \ref{fig:vtab_num_adapter_dev} show how the threshold influences the number of added adapters (at each layer), displaying trends consistent with those in Fig. \ref{fig:ina_acc_dev} and \ref{fig:vtab_acc_dev}. Fig. \ref{fig:ina_acc_dev} and \ref{fig:ina_num_adapter_dev} show that a smaller expansion threshold leads to more frequent expansion, which could boost the performance at some level through more parameters. A threshold that is too large (\eg, values over 1.5) minimizes the chance for expansion, which may lead to insufficient adaptation. In SEMA, a proper expansion threshold within a wide range can lead to a balance between the performance gain and the parameter size.

\begin{figure}[htb]
\vspace{-0.3cm}
  \centering
  \begin{subfigure}[b]{0.45\linewidth}
    \includegraphics[width=\linewidth]{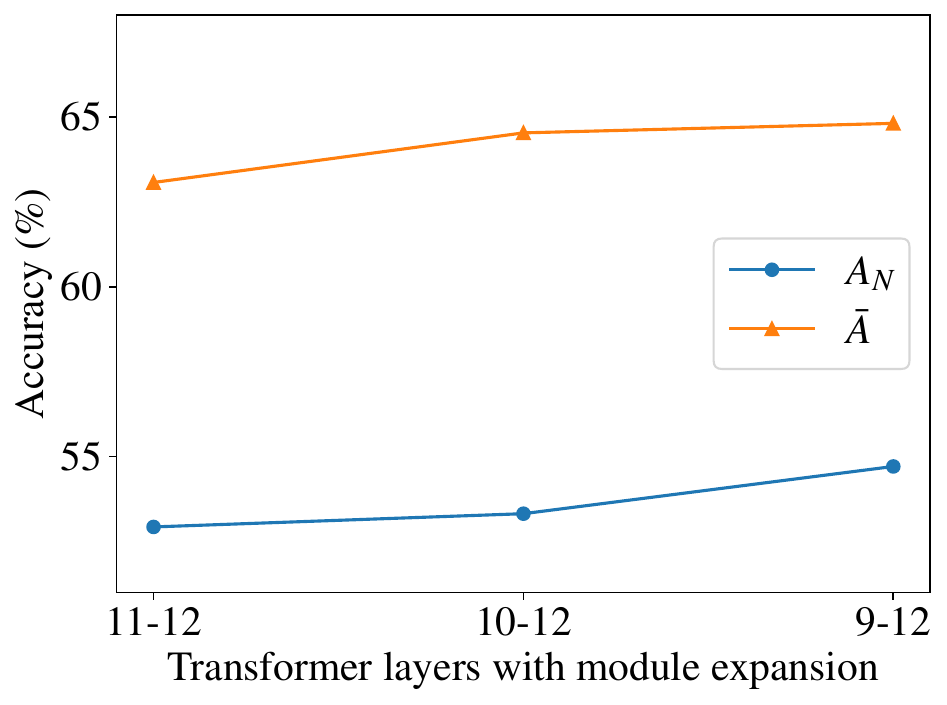}
    \caption{Accuracy}
    \label{fig:ina_acc_n_layer}
  \end{subfigure}
  \hfill
  \begin{subfigure}[b]{0.45\linewidth}
    \includegraphics[width=\linewidth]{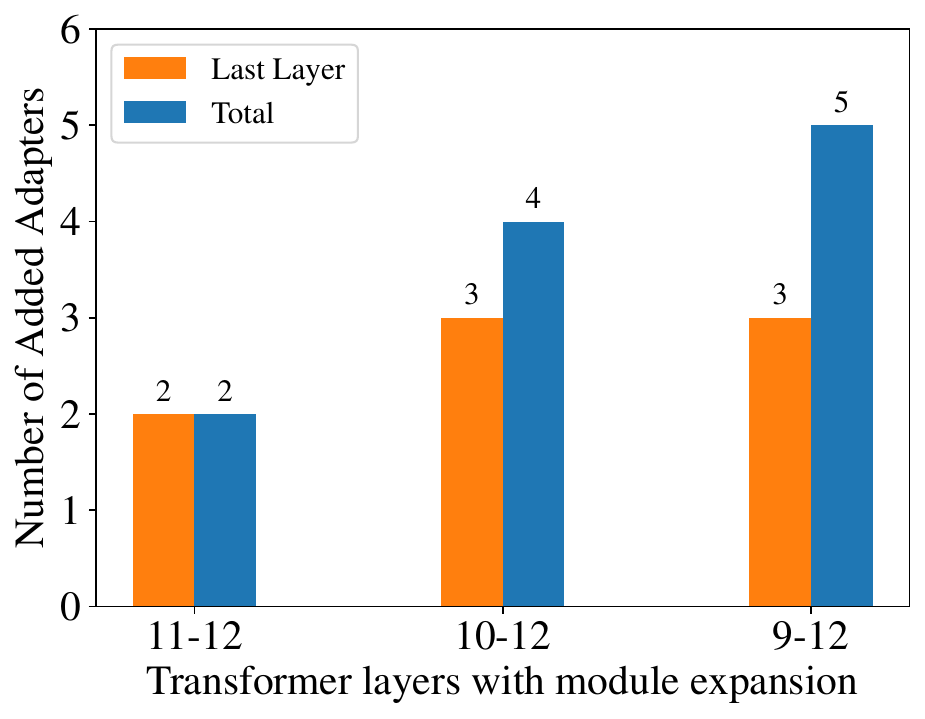}
    \caption{Num. of adapters}
    \label{fig:ina_num_adapter_n_layer}
  \end{subfigure}
  \vskip0.5\baselineskip
   \begin{subfigure}[b]{0.45\linewidth}
    \includegraphics[width=\linewidth]{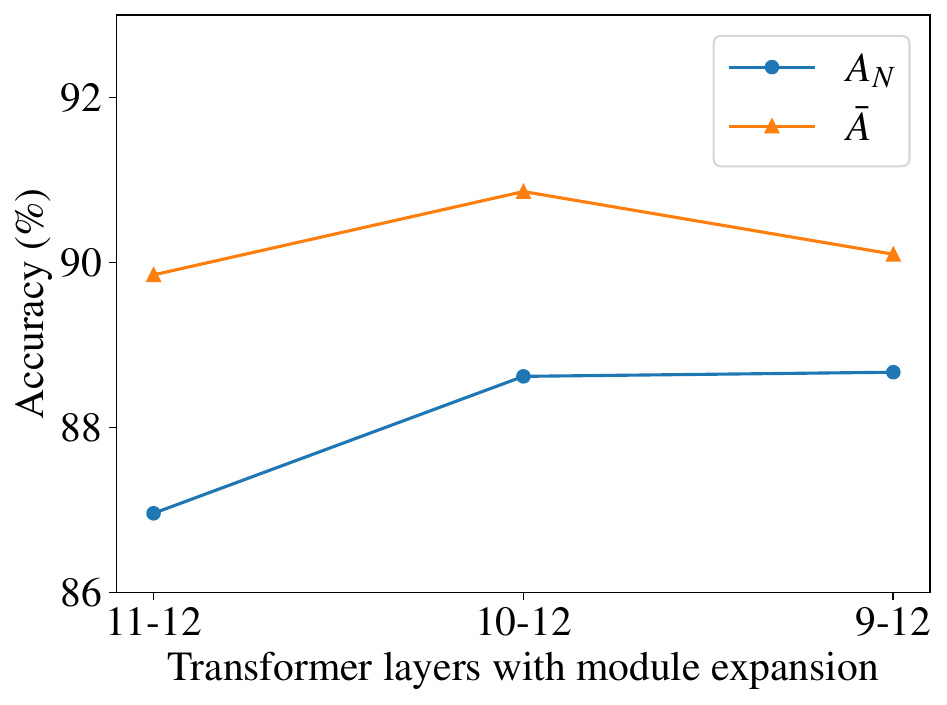}
    \caption{Accuracy}
    \label{fig:vtab_acc_n_layer}
  \end{subfigure}
  \hfill
  \begin{subfigure}[b]{0.45\linewidth}
    \includegraphics[width=\linewidth]{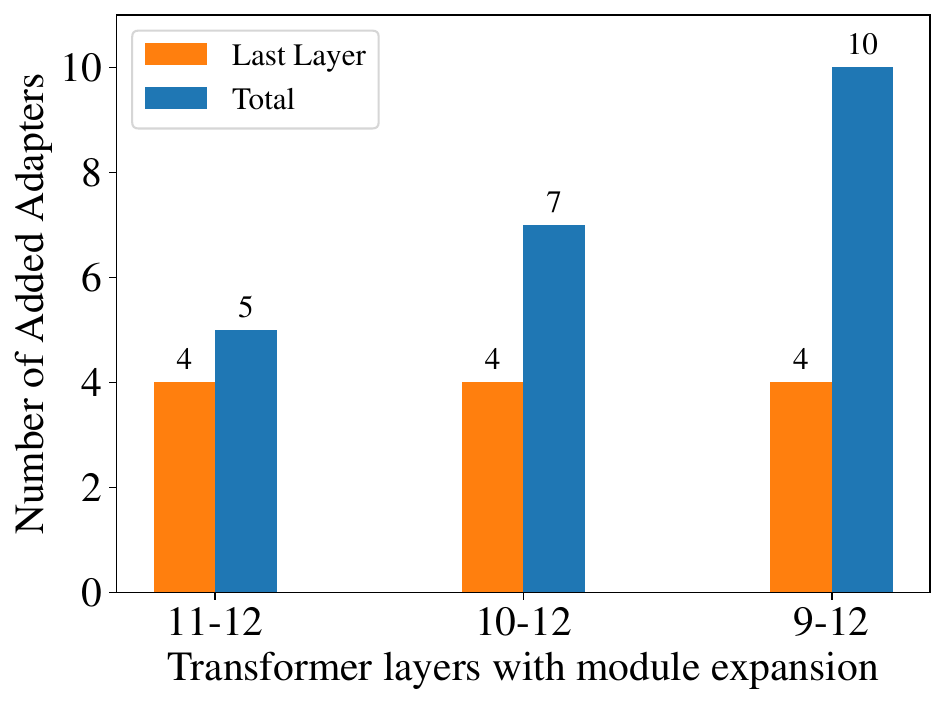}
    \caption{Num. of adapters}
    \label{fig:vtab_num_adapter_n_layer}
  \end{subfigure}
  \vspace{-0.15cm}
  \caption{Analysis of the effect of multi-layer expansion, with (a)(b) ImageNet-A and (c)(d) VTAB. By enabling automatic self-expansion on multiple transformer layers, SEMA can achieve better performance than restricting that on a single layer.}
  \label{fig:vtab_n_layer}
  \vspace{-0.3cm}
\end{figure}

\noindent\textbf{Analysis of multi-layer expansion.}
In Fig. \ref{fig:vtab_n_layer}, we explore the effects on accuracy by implementing expansion across varying numbers of layers, ranging from the last 2 layers (\#11-\#12) to the last 4 layers (\#9-\#12). Intuitively, allowing expansion in deeper layers enables better adaptation to different tasks. However, as shown in Fig. \ref{fig:ina_num_adapter_n_layer} and Fig. \ref{fig:vtab_num_adapter_n_layer}, permitting expansion in early transformer layers also increases the overall number of added adapters, without a significant boost in performance as earlier layers tend to behave similarly despite distribution shifts. Also, enforcing addition of too many adapters may cause difficulty in training, especially in early transformer layers.

\begin{table}[h]
\centering
 
\small
\begin{tabular}{@{}lccccccccc cccccccc}
    \toprule
    \multicolumn{1}{l}{{Method}} & 
    \multicolumn{2}{c}{ImageNet-A}  & \multicolumn{2}{c}{VTAB}  \\

 & ${\bar{\mathcal{A}}}$ & {$\mathcal{A}_N $}  & ${\bar{\mathcal{A}}}$ & {$\mathcal{A}_N $}
\\
\midrule
Adapter\cite{adaptformer}  & \textbf{64.53} & \textbf{53.32} & 91.26 & \textbf{89.64}  \\
    \midrule
    LoRA\cite{hu2021lora}    & 63.50 & 52.67 & \textbf{91.85} & 88.53 \\
Convpass\cite{jie2022convpass}  &  63.48 & 51.74  & 90.68 &  88.62  \\
    \bottomrule
\end{tabular}
\vspace{-0.2cm}
    \caption{
 Different adapter variants.
 }
     \label{tab:adapter_ablation}
\end{table}

\begin{figure}[htp]
    \centering
    \includegraphics[width=0.6\linewidth]{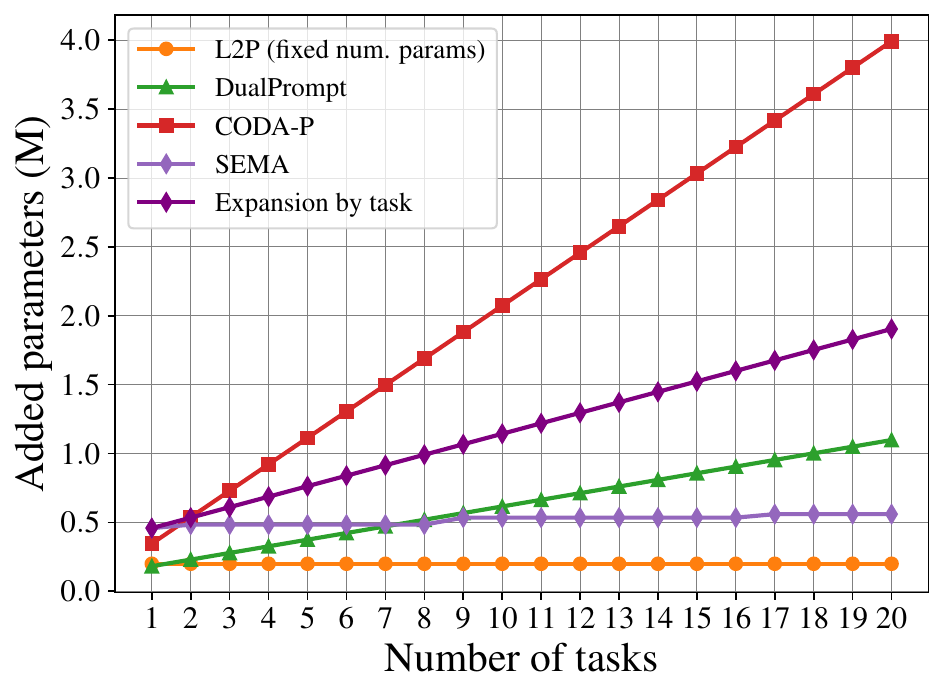}
    \vspace{-0.3cm}
    \caption{Analysis on added parameters (in Millions) during model deployment on ImageNet-A. }
    \label{fig:param_trend}
    \vspace{-0.3cm}
\end{figure}

\noindent\textbf{Ablation studies on adapter variants.} 
 Apart from Adapter \cite{adaptformer}, we extend our evaluation to other 
variants, namely LoRA \cite{hu2021lora} and Convpass \cite{jie2022convpass}. As shown in Tab. \ref{tab:adapter_ablation}, our proposed approach is robust to the choice of adapter methods, showing the broad applicability and effectiveness of our dynamic expansion strategy across different adapter methods.

\noindent\textbf{Sub-linear growth of parameters.} 
In Fig. \ref{fig:param_trend}, instead of expanding w.r.t. number of tasks, SEMA adds parameters at a sub-linear rate, showing the efficiency of the self-expansion mechanism. Further analysis is provided in Appendix \ref{supp:sub-linear_param_growth}.

\section{Conclusion}
\label{sec:conclusion}
In this paper, we propose a novel self-expandable modularized adaptation approach for continual learning. SEMA learns to reuse and add modules in an automated manner without memory rehearsal. We incorporate an efficient expansion strategy with detection for feature distribution shifts in different layers of transformer-based models, successfully mitigating the forgetting problem of jointly using the fixed set of parameters. Experimental results demonstrate the outstanding performance of SEMA to datasets with different levels of distribution shifts.

\noindent\textbf{Limitations and future work.} 
We perform the task-oriented expansion at most once per layer for each task considering the CIL characteristics and parameter efficiency. The design can be more flexible to enable fully online dynamic expansion, which could open possibility in better adaptation for data with intra-task diversity and enable online CL. Moreover, the expansion of SEMA is based on the distribution shift detection ability from RDs, which could be further enhanced by elevating the optimization of RDs and expansion protocol to a meta level with a closed loop.

{
    \small
    \bibliographystyle{ieeenat_fullname}
    \bibliography{main}
}
\clearpage

\setcounter{table}{3}
\setcounter{figure}{8}

\setcounter{page}{1}
\maketitlesupplementary

\appendix

\section{More Details about SEMA}
\subsection{More Details of SEMA Training}
\label{supp:training_procedure}
\begin{figure*}[!h]
    \centering
    \includegraphics[width=\linewidth]{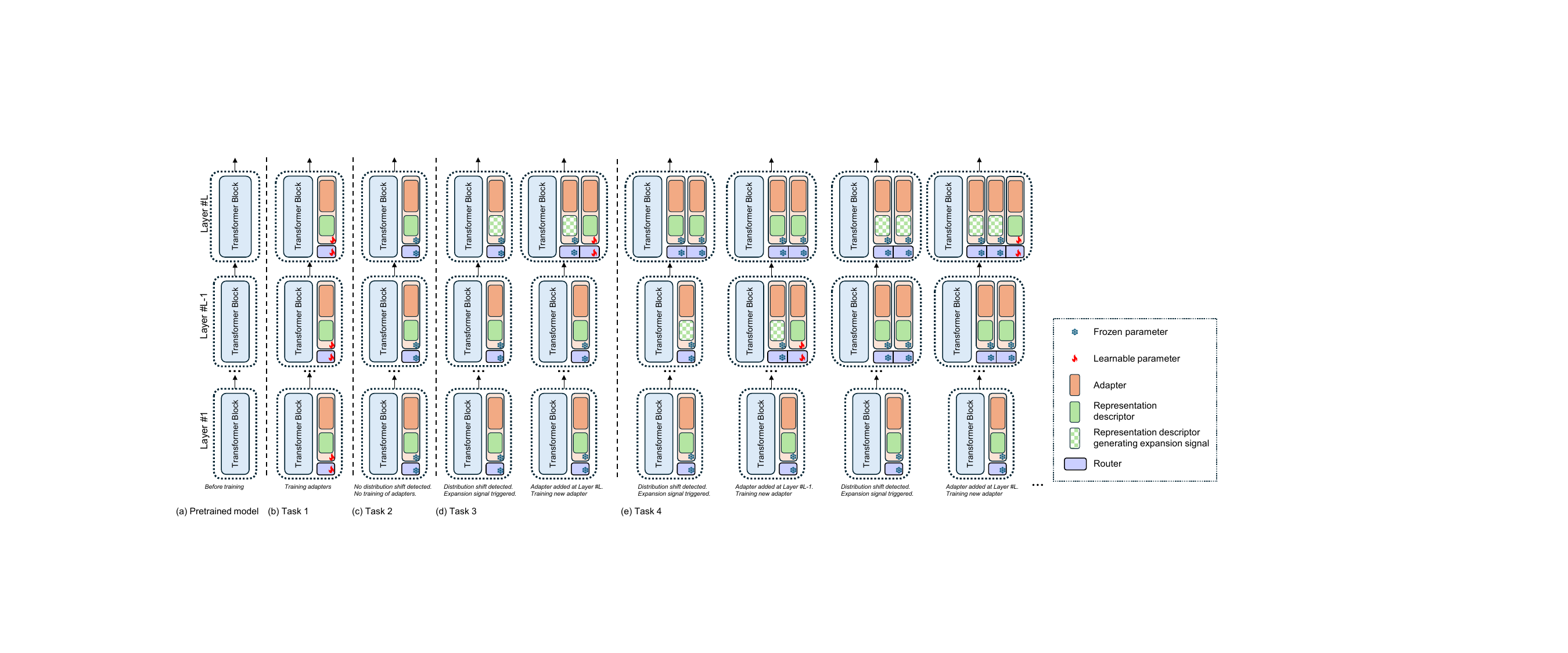}
    \caption{A more detailed example for the illustration of the learning process. (a) The pre-trained model with $L$ transformer layers is provided for adaptation. (b) At the start of training, each transformer layer is equipped with one expandable weighting router and one adapter module, including one functional adapter and its paired representation descriptor. All modules are trainable at this stage. (c) All modules and routers are frozen after the training on Task 1. When Task 2 arrives, the detection of distribution shift is performed with all frozen representation descriptors in each transformer layer for all batches in Task 2. Since no distribution shift is observed, module addition is not performed and all modules are frozen. (d) As Task 3 arrives, the detection for the distribution shift is executed again and the distribution shift is observed in the $L$-th layer. Expansion signal is triggered and an adapter module is added in the $L$-th layer with the expanded router. Training for the newly added adapter and router is performed. Since the addition is performed at the last transformer layer, no further detection for distribution shift is required. (e) When Task 4 arrives, expansion signal is triggered in the $L-1$-th layer during the detection phase. After sufficient training, the newly added module is frozen and detection for distribution shift in later layers is executed. When both representation descriptors in the $L$-th layer consider the incoming feature as an outlier, expansion signal will be triggered. A new module is added for training in the $L$-th layer while all other modules are frozen.}
    \label{fig:supp_process}
\end{figure*}

We discuss more details of SEMA training using a more detailed example in Fig. \ref{fig:supp_process}, which contains more details (\ie, different types of cases and the distribution shift detection/scanning procedure) compared to that in Fig. \ref{fig:process}. 
At the start of the training, each transformer block at different layers is equipped with one adapter module containing one adapter and one representation descriptor, as well as an expandable weighting router, as shown in Fig. \ref{fig:supp_process} (b). They are added as the default adapters and trained on the first task. 
After the first task, for the incoming new tasks, SEMA monitors the representations of each batch of samples at each layer with the AE-based representation descriptor. As discussed in Sec. \ref{sec:expansion_strategy}, the distribution shift is measured using the $z$-score computed from the mean and standard deviation of reconstruction errors stored in a buffer. This buffer is implemented as a fixed stack of 500 samples, maintaining reconstruction errors from the most recent batches. New adapters are added if a significant enough representation/distribution shift is detected at each layer. Adding the adapters expands the model's representation ability for handling the new patterns. As introduced in the main paper, SEMA performs task-oriented expansion (in the class-incremental learning setting given the task boundary in training), adding at most one adapter per layer. 
As shown in Fig. \ref{fig:process} and Fig. \ref{fig:supp_process}, the detection and expansion operation starts from the transformer layers closest to the input. Once a significant distribution shift is detected at a specific layer that could not be handled by \emph{all} existing adapters (detected by RDs), an expansion signal is triggered in this layer/block. A new adapter module will be added to the layer where the expansion signal is triggered, along with an expansion of the weighting router, and activated for training. After sufficient training, the detection phase will be restarted for the later layers. If no distribution shift is reported for a task in any layers, as shown in Fig. \ref{fig:supp_process} (c), no adapter module will be added, and no training of adapters is required for this task. 

\section{More Details about Implementation and Evaluation}
\label{supp:exp}
\subsection{Details of Datasets}

\textbf{CIFAR-100} contains 100 classes with 500 training samples and 100 testing samples per class. \\
\textbf{ImageNet-R} contains renditions of 200 ImageNet classes, which is a challenging CL benchmark introduced by with great intra-class diversity.\\ 
\textbf{ImageNet-A} contains real-world images filtered from ImageNet in an adversarial manner which are hard to be classified by models pre-trained with ImageNet.   \\
\textbf{VTAB} consists of 50 classes from 5 domains with 10 classes from each domain. 

To construct class-incremental setting, for results reported in Tab. \ref{tab:cil_result}, CIFAR-100, ImageNet-A and VTAB are split in a manner where each task consists of 10 distinct classes. ImageNet-R is reported with results for 5 tasks (40 classes per task), 10 tasks (20 classes per task), and 20 tasks (10 classes per task).

\subsection{Implementations of Compared Methods}
For SimpleCIL and ADAM, we use the official implementation at \url{https://github.com/zhoudw-zdw/RevisitingCIL}. For InfLoRA, we use the official implementation at \url{https://github.com/liangyanshuo/InfLoRA}. For other prompting methods, namely L2P, DualPrompt and CODA-P, we adopt the open-source implementation from PILOT toolbox~\cite{sun2023pilot}, available at \url{https://github.com/sun-hailong/LAMDA-PILOT}. In our experiments, we adhere to the hyperparameter configurations as specified in the original publications for each of the compared methods, We use ViT-B/16-IN1K as the backbone with the same data shuffling as \cite{revisiting} for all methods.

\subsection{Details on Evaluation Metrics}

Denote the accuracy of the $i$-th task after training on the $N$-th task as $\mathcal{A}_{i, N}$. The average accuracy $\mathcal{A}_N$ represents the average accuracy of all seen tasks after training on the $N$-th task:
$$
\mathcal{A}_N = \frac{1}{N} \sum_{i=1}^{N}\mathcal{A}_{i, N},
$$
which is often considered as the most important evaluation metric in continual learning.

The average incremental accuracy $\bar{\mathcal{A}}$ is the average accuracy along incremental stages, defined as:
$$
\bar{\mathcal{A}} = \frac{1}{N}\sum_{t=1}^{N}\mathcal{A}_{t}.
$$

\begin{table*}[htp]
\centering
    \begin{tabular}{@{}lccccccccc cccccccc}
        \toprule
        \multicolumn{1}{l}{{Method}} & 
        \multicolumn{2}{c}{CIFAR-100}
            & \multicolumn{2}{c}{5-Task IN-R}
            & \multicolumn{2}{c}{10-Task IN-R}
            & \multicolumn{2}{c}{20-Task IN-R} 
        & \multicolumn{2}{c}{ImageNet-A} & \multicolumn{2}{c}{VTAB} \\
		 & ${\bar{\mathcal{A}}}$ & {$\mathcal{A}_N $}
         & ${\bar{\mathcal{A}}}$ & {$\mathcal{A}_N $}
       & ${\bar{\mathcal{A}}}$  & {$\mathcal{A}_N $} 
        & ${\bar{\mathcal{A}}}$ & {$\mathcal{A}_N $}
       & ${\bar{\mathcal{A}}}$  & {$\mathcal{A}_N $} 
       & ${\bar{\mathcal{A}}}$   & {$\mathcal{A}_N $} 
    \\
    \midrule
L2P                      & 89.51          & 85.02          & 72.90             & 65.83          & 74.55             & 69.75          & 74.49              & 65.82             & 46.67             & 39.30          & 79.17          & 63.56          \\
DualPrompt               & 90.39          & 85.64          & 73.91             & 68.81          & 73.10              & 67.18          & 73.67              & 68.88             & 58.45             & 48.78          & 88.11          & 77.58          \\
CODA-P                   & 91.01          & 86.20          & 79.78             & 74.68          & 79.15              & 73.05          & 70.36              & 65.32             & 50.73             & 37.06          & 85.13          & 85.85          \\
SimpleCIL                & 87.13          & 81.26          & 59.70             & 54.33          & 61.12              & 54.33          & 61.92              & 54.33             & 60.50             & 49.44          & 85.99          & 84.38          \\
ADAM                     & 92.18          & 87.47          & 77.28             & 70.58          & 76.71              & 69.18          & 75.08              & 67.30             & 60.53             & 49.57          & 85.95          & 84.35          \\
InfLoRA                  & 91.71          & 86.73          & 81.75             & 76.77          & 81.38              & 74.72          & 76.97              & \textbf{69.65}    & 56.84             & 41.61          & 89.61          & 86.52          \\
\midrule
SEMA                     & \textbf{92.23} & \textbf{87.84} & \textbf{83.27}    & \textbf{77.13} & \textbf{81.39}     & \textbf{74.82} & \textbf{77.84}     & 69.60             & \textbf{62.50}    & \textbf{51.35} & \textbf{91.99} & \textbf{90.86} \\
        \bottomrule
\end{tabular}
 \caption{Experiments on class-incremental learning benchmarks with ViT-B/16-IN21K weight.}
 \label{tab:21k_result}	
\end{table*}

\section{More Experiments and Ablation Studies}
\subsection{Influence of Pre-trained Weights}
\label{supp:21k_weight}
In the main paper, we experiment SEMA and other methods with ViT-B/16-IN1K in Tab. \ref{tab:cil_result}. To study the influence of pre-trained weights, we further experiment SEMA with another commonly used pre-trained ViT weight, i.e., ViT-B/16-IN21K. We evaluate the performance using average accuracy $\mathcal{A}_N$ and average incremental accuracy $\bar{\mathcal{A}}$. As shown in Tab. \ref{tab:21k_result}, SEMA consistently outperforms prompting and adaptation methods in most class-incremental learning settings. This indicates that our model is robust in performance regardless of different choices of pre-trained weights.

\begin{table}[!t]
\centering
\resizebox{\linewidth}{!}{
\begin{tabular}{lcccc}
\toprule
\multirow{2}{*}{Dataset} & \multicolumn{2}{c}{Expansion by Task} & \multicolumn{2}{c}{SEMA}      \\
                         & Params (M)      & $\mathcal{A}_N $     & Params (M) & $\mathcal{A}_N $ \\
                        \midrule
CIFAR-100                & 1.066           & 86.86                & 0.645      & 86.98            \\
ImageNet-R               & 1.904           & 74.08                & 0.617      & 74.53            \\
ImageNet-A               & 1.904           & 52.80                 & 0.560      & 53.32            \\
VTAB                     & 0.647           & 89.09                & 0.554      & 89.64           \\
\bottomrule
\end{tabular}
}
\caption{Comparison of added parameters and accuracy with different expansion strategies. 
``Expansion by Task'' is a \emph{naive} implementation of SEMA's variant that adds one set of adapters (at all layers allowing expansion) for every new task. SEMA only expands if a distribution shift is detected by the representation descriptor.}
\label{tab:per-task_expansion}
\end{table}

\subsection{Further Analyses on the Effectiveness of Self-Expansion}
\label{supp:sub-linear_param_growth}
The proposed method SEMA enables the model to add parameters and expand its capacity on demand. It allows the model to handle samples that could not be handled before by adding a small number of parameters. In continual learning, this process helps to alleviate forgetting by avoiding interference from new patterns while still encouraging knowledge reuse and transfer. 
Unlike some methods \cite{coda,dualprompt,ease} that continually adding task-specific modules by task with a \emph{linear} parameter growth rate, SEMA produces a \emph{sub-linear} expansion rate, w.r.t. number of seen tasks. 
To analyze and show the effectiveness of this self-expansion process, we conducted comparisons on four different settings where CIFAR-100, ImageNet-R, ImageNet-A and VTAB contain 10 tasks, 20 tasks, 20 tasks and 5 tasks respectively, corresponding to four settings reported in Fig. \ref{fig:acc_record}. We compare with other related methods and a \emph{naive implementation} of the ``expansion-by-task'' variant of SEMA. This simple variant model incrementally adds adapters to the layers that allow expansion for each incoming task. 
The number of parameters and accuracy are reported in Tab. \ref{tab:per-task_expansion}. 
Despite the naive implementation of ``expansion-by-task'', the results in Tab. \ref{tab:per-task_expansion} show that SEMA with flexible self-expansion can achieve better performance than that using more parameters. We demonstrate that our expansion strategy is efficient in both controlling the size of added parameters, regardless of the length of task sequence, encouraging knowledge reuse and reducing potential task interference in adapter weighting.

\begin{table*}[htp]
\resizebox{\linewidth}{!}{
\begin{tabular}{@{}lccccccccc@{}}
\toprule
\multirow{2}{*}{Type} & \multirow{2}{*}{Method} & \multicolumn{2}{c}{CIFAR-100} & \multicolumn{2}{c}{ImageNet-R} & \multicolumn{2}{c}{ImageNet-A} & \multicolumn{2}{c}{VTAB} \\
 & & \multicolumn{1}{l}{Params (M)} & $\mathcal{A}_N $ & \multicolumn{1}{l}{Params (M)} & $\mathcal{A}_N $ & \multicolumn{1}{l}{Params (M)} & $\mathcal{A}_N $ & \multicolumn{1}{l}{Params (M)} & $\mathcal{A}_N $ \\ \midrule
Fixed Param Size & L2P & 0.123 & 77.87 & 0.200 & 62.90 & 0.200 & 38.48 & 0.085 & 80.83 \\
\midrule
\multirow{3}{*}{Expandable Param Size} & DualPrompt & 1.022 & 80.43 & 1.098 & 61.97 & 1.098 & 50.23 & 0.983 & 79.79 \\
& CODA-P & 3.917 & 86.11 & 3.994 & 70.02 & 3.994 & 35.02 & 3.878 & 81.58 \\ 
& SEMA & \textbf{0.645} & \textbf{86.98} & \textbf{0.617} & \textbf{74.53} & \textbf{0.560} & \textbf{53.32} & \textbf{0.554} & \textbf{89.64} \\ \bottomrule
\end{tabular}
}
\caption{Number of added parameters used in model deployment, measured in Millions. L2P uses a fixed size of prompts. DualPrompt and CODA-P incrementally add parameters (\ie, prompts) sequentially by task. SEMA adds a small number of parameters with its dynamic expansion strategy.}
\label{tab:param_size}
\end{table*}

\begin{figure}[!t]
    \centering
    \includegraphics[width=0.45\textwidth]{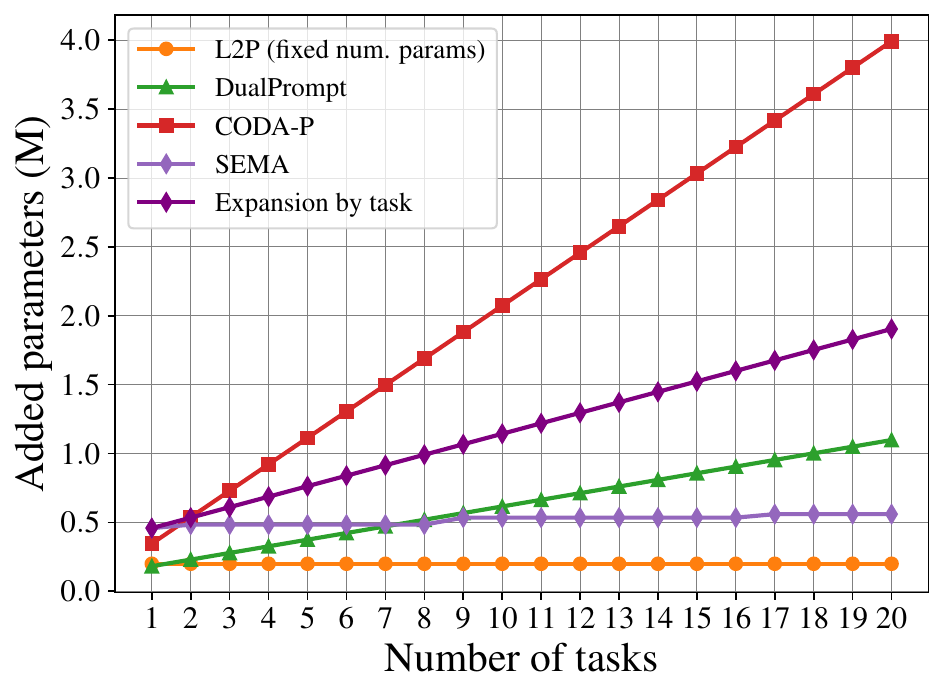}
    \caption{Analysis on added parameters (in Millions) during model deployment on ImageNet-A. We compare with methods using fixed number of prompts like L2P, and methods like DualPrompt and CODA-P that incrementally expand like SEMA but with prompts and on a linear basis according to tasks. Expansion by task adds adapters for every incoming task, whilst SEMA executes expansion on demand, which increments parameters on a sub-linear basis. Specifically, SEMA added more parameters (with expansions at more layers) at Task 9 than other steps with expansion. }
    \label{supp:param_trend}
\end{figure}

Tab. \ref{tab:param_size} reports the size of added parameters in several different PTM-based methods. 
While L2P uses a fixed size of prompt pool with small amount of added parameters, the fixed size of trainable parameters may limit its capability to adapt to more distribution shifts in continual learning and comes with a higher chance of forgetting. 
Compared to other methods (\ie, CODA-P and DualPrompt) that incrementally add parameters (\ie, prompts in these methods) for each task, SEMA involves much fewer added parameters in the model. Apart from the adaptation approach and expansion strategy, the compared methods in this part use similar techniques as the proposed method (such as the classifier and PTMs). Note that the added parameters for SEMA only consider the functional adapters that are used in deployment. The RDs are maintained for training and updating of the model, which can be handled in parallel to other parameters and do not influence the deployment of the model. 
As shown in Fig. \ref{supp:param_trend} (also demonstrated in the main paper \cref{fig:param_trend}), SEMA can dynamically expand the model with a small \emph{sub-linear} rate, while the other methods are usually with a \emph{linear} rate.  

\subsection{Further Discussions on the Weighting Router}
\label{supp:router_rd}
\noindent\textbf{Routing relying on representation descriptor.}
In SEMA, we use the representation descriptors (RDs) to capture the distribution of the input representations corresponding to each modular adapter, which are used to detect novel patterns triggering the expansion signal. 
The RDs can be used to compose the adapters via hard selection, as in similar modular networks. Specifically, the reconstruction error of the AE-based RDs can provide the identity information of each inference sample, w.r.t. the adapters, at different layers. However, the RD-based adapter selection/routing can be unreliable for every single individual input, and related works usually rely on the statistics of a batch of samples \cite{lmc}, limiting the application. We thus propose directly learning the soft weighting router for mixture usage of the adapters. 
To analyze the behavior of the RDs in detail, we conduct the experiments that perform adapter composing relying on the RDs and show the results in Tab. \ref{tab:ae_routing}. As shown in Tab. \ref{tab:ae_routing}, the RD-based routing can achieve sound performance on most datasets, which validates the representation ability of RDs. SEMA with the soft weighting router can perform better, relying on the specifically learned router that is trained together with the adapters.

\begin{table*}[ht]
	\centering
		\begin{tabular}{@{}lccccccccc cccccccc}
			\toprule
			\multicolumn{1}{l}{{Method}} & 
			\multicolumn{2}{c}{CIFAR-100}
            & \multicolumn{2}{c}{5-Task IN-R}
            & \multicolumn{2}{c}{10-Task IN-R}
            & \multicolumn{2}{c}{ 20-Task IN-R} 
        & \multicolumn{2}{c}{ImageNet-A} & \multicolumn{2}{c}{VTAB} \\
		 & ${\bar{\mathcal{A}}}$ & {$\mathcal{A}_N $}
         & ${\bar{\mathcal{A}}}$ & {$\mathcal{A}_N $}
       & ${\bar{\mathcal{A}}}$  & {$\mathcal{A}_N $} 
       & ${\bar{\mathcal{A}}}$ & {$\mathcal{A}_N $}
       & ${\bar{\mathcal{A}}}$  & {$\mathcal{A}_N $} 
       & ${\bar{\mathcal{A}}}$   & {$\mathcal{A}_N $} 
		\\
\midrule
SEMA             & \textbf{91.37} & \textbf{86.98}  & \textbf{84.75} & \textbf{79.78} & \textbf{83.56} & \textbf{78.00} & \textbf{81.75} & \textbf{74.53}   & \textbf{64.53} & \textbf{53.32}   & \textbf{91.26} & \textbf{89.64} \\
\midrule
RD-based routing & 90.91          & 83.61           & 84.46          & 79.50          & 82.76          & 76.63          & 81.02          & 74.13            & 61.80          & 50.36            & 90.83          & 88.53         \\
\bottomrule
    \end{tabular}
    \caption{Comparison between routing with the expandable weighting router and RD-based routing.}
     \label{tab:ae_routing}		
\end{table*}

\begin{figure*}[!t]
    \centering
    \includegraphics[width=0.45\textwidth]{images/router_weighting.pdf} \\
    \includegraphics[width=0.45\textwidth]{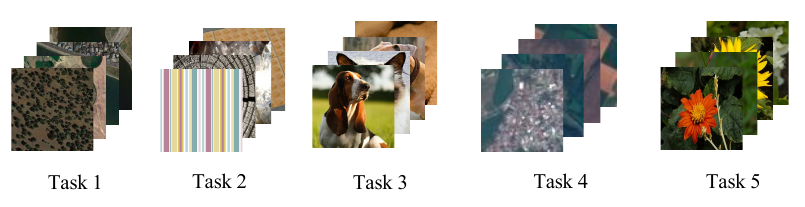}
    \caption{Adapter usage visualization on VTAB (same as Fig. \ref{fig:adapter_routing}). For clear and simplified visualization, we only allow expansion at the last transformer layer. We report the average adapter usage of each task. We also provide visual illustrations of sample images from each VTAB task.}
    \label{supp:adapt_weight}
\end{figure*}

\begin{table*}[!t]
\centering
\begin{tabular}{@{}lcccc@{}}
\toprule
\multirow{2}{*}{Method} & \multicolumn{4}{c}{Train Time (s)}                                                                                \\
                        & \multicolumn{1}{l}{CIFAR-100} & \multicolumn{1}{l}{ImageNet-R} & \multicolumn{1}{l}{ImageNet-A} & \multicolumn{1}{l}{VTAB} \\ \midrule
L2P          & 0.27      & 0.27       & 0.29       & 0.28 \\
DualPrompt    & \textbf{0.25}      & 0.25       & 0.27       & \textbf{0.29} \\
CODA-P         & 0.31      & 0.32       & 0.35       & 0.36 \\
\midrule
SEMA (Overall) & \textbf{0.25}      & \textbf{0.11 }      & \textbf{0.15}       & 0.31 \\
- Adapter     & 0.13      & 0.10       & 0.12       & 0.20 \\
- RD         & 0.12      & 0.01       & 0.03       & 0.11 \\
\bottomrule
\end{tabular}
\caption{Average per-batch train time of each method on each task measured in seconds. SEMA (overall) denotes the training time used when adapter and representation descriptor (RD) are trained sequentially.}
\label{tab:train_time}
\end{table*}

\begin{table*}[!t]
\centering
\begin{tabular}{@{}lcccc@{}}
\toprule
\multirow{2}{*}{Method} & \multicolumn{4}{c}{Inference Time (ms)}                                                                                \\
                        & \multicolumn{1}{l}{CIFAR-100} & \multicolumn{1}{l}{ImageNet-R} & \multicolumn{1}{l}{ImageNet-A} & \multicolumn{1}{l}{VTAB} \\ \midrule
L2P                    &  9.44    & 9.53 & 9.86 & 9.46           \\
DualPrompt              & 9.44    & 9.51 & 9.84 & 9.44  \\
CODA-P                  &  9.45  & 9.47 & 9.85  & 9.43  \\
ADAM                    & 9.95   & 10.03  &  10.36  & 9.45  \\ \midrule
SEMA                    &  \textbf{4.48}  &  \textbf{7.39} & \textbf{9.01}  &  \textbf{7.38} \\ \bottomrule
\end{tabular}
\caption{Per-image inference time of each method measured in milliseconds.}
\label{tab:inf_time}
\end{table*}

\noindent\textbf{More discussions on adapter usage.} Fig. \ref{fig:adapter_routing} shows the average adapter usage of each task on VTAB. For clear visualization, we enable expansion to be performed only at the last layer and attach sample images from each task in Fig. \ref{fig:adapter_routing}. Adapter 1, Adapter 2, and Adapter 3 are automatically added and trained when Task 1, Task 2, and Task 3 arrive, respectively. Task 1, Task 2, and Task 3 all present high preference for choosing the adapters that were trained with them, showing the effectiveness of the router to direct samples to the adapter that is trained with a similar distribution. 
While adapter expansion is not triggered for Task 4, Task 4 data largely employs Adapter 1 during inference. 
As visualized in Fig. \ref{supp:adapt_weight}, the data distribution between Task 1 (remote sensing images) and Task 4 (land cover) is similar. 
Similarly, Task 3 (pets) and Task 5 (flowers) both comprise natural images with similar characteristics, hence have higher similarity in distribution than Task 1 (remote sensing images) and Task 2 (texture images), and exhibit a preference for Adapter 3. Thus, we show that our expandable weighting router can effectively select the proper mixture pattern of adapters with various data distributions.

\subsection{Training and Inference Time}
\label{supp:train_time}

All experiments can be produced on a single NVIDIA GeForce RTX 3090 GPU. To compare the training efficiency, we report the per-batch training time averaged over the incremental learning process in Tab. \ref{tab:train_time}.
Similar to Tab. \ref{tab:per-task_expansion}, ImageNet-R here is split into 20 tasks with 10 classes per task. Note that the training processes of adapter and representation descriptor in each adapter module of SEMA are in parallel after expansion, thus the training of these two components can be performed in parallel with multiple GPUs. 
We report the training time of adapters (\ie, ``Adapter'' in Tab. \ref{tab:train_time}) and representation descriptors (\ie, ``RD'' in Tab. \ref{tab:train_time}) separately, along with the overall time usage of SEMA training if adapters and representation descriptors are trained sequentially. 

SEMA with components trained in a parallel manner is highly efficient. Even without the parallel setup, training the adapters and RDs in SEMA in sequence can still be faster than other PTM-based CL methods on most datasets. As SEMA only expands while encountering distribution shifts in incoming new tasks, for tasks that do not trigger expansion, no training of adapters and representation descriptors is performed and training time on these tasks is minimized, leading to training efficiency in the long term. Note that the scanning for distribution shifts is stopped as long as a batch of data triggers expansion behaviour, which is more efficient comparing to InfLoRA which requires processing through all data in the given task twice for LoRA initialization before training and post-training computation for gradient projection memory.

\begin{figure*}[!t]
  \centering
  \begin{subfigure}[b]{0.31\linewidth}
    \includegraphics[width=\linewidth]{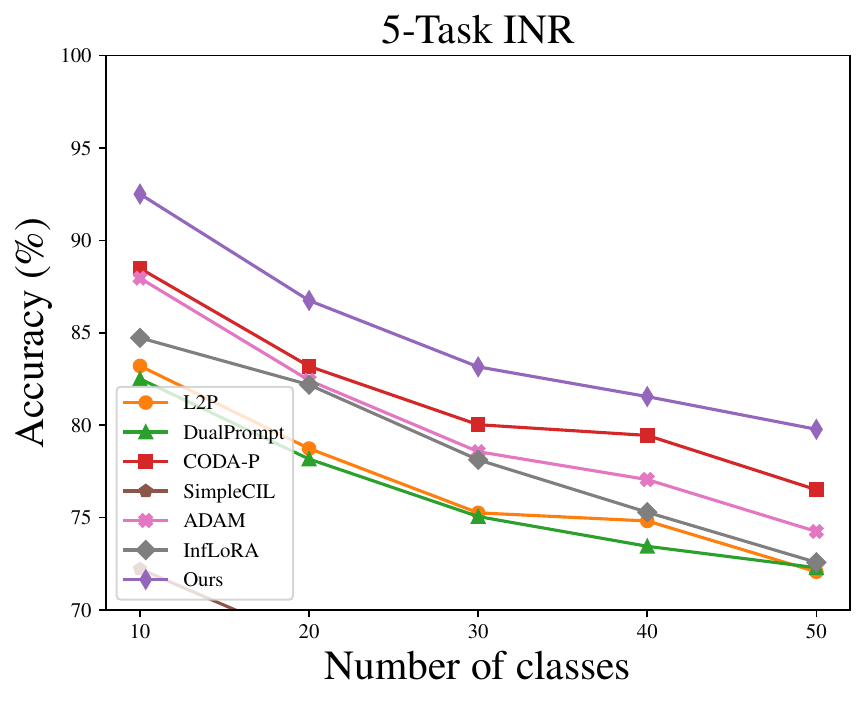}
  \end{subfigure}
  \begin{subfigure}[b]{0.31\linewidth}
    \includegraphics[width=\linewidth]{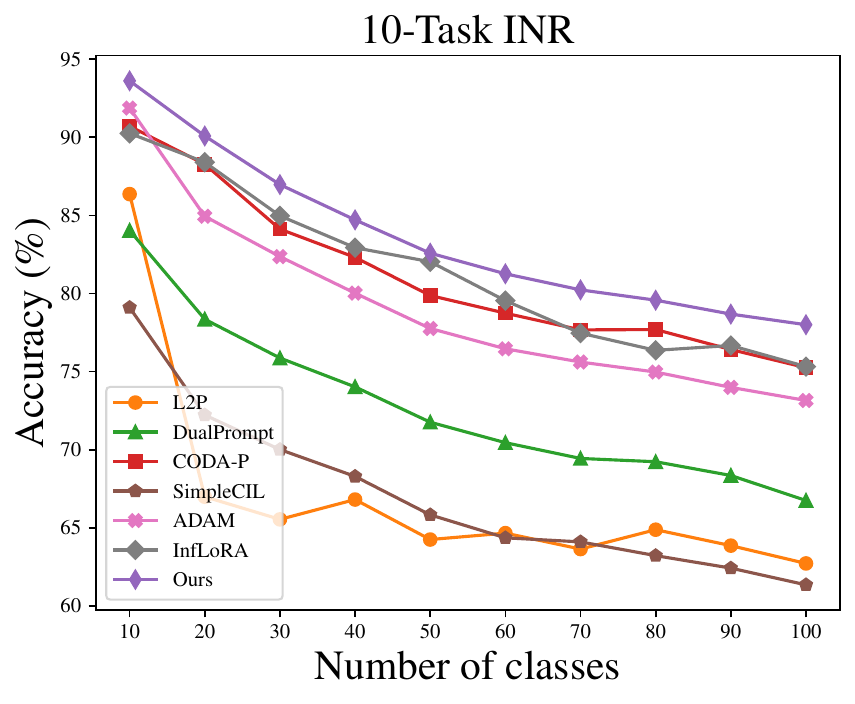}
  \end{subfigure}
  \begin{subfigure}[b]{0.31\linewidth}
    \includegraphics[width=\linewidth]{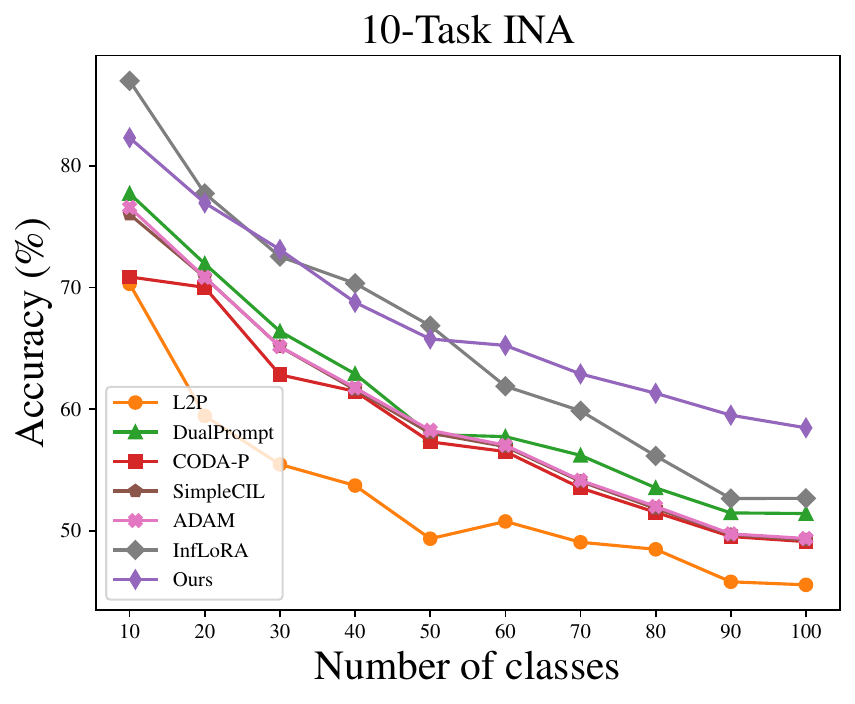}
  \end{subfigure}
  \caption{More results on incremental performance for ImageNet-R and ImageNet-A.}
  \label{fig:acc_record_supp}
\end{figure*}

We evaluate the inference efficiency and report the average inference time of each image measured in milliseconds in Tab. \ref{tab:inf_time}. We show that SEMA is efficient compared to other methods on all datasets. 
The inference latency of the listed prompting continual learning methods is caused by the extra procedure of processing the image with a frozen pre-trained model for the query function. Similarly, ADAM requires extra feature extraction with a frozen pre-trained model for the concatenation of pre-trained features and adapted features. SEMA relieves the dependency on the frozen pre-trained model as we focus on the intermediate feature distribution of each transformer block.

\subsection{Additional Results on Longer Task Sequence}
We perform the 50-step experiment on ImageNet-R and ImageNet-A, where each task contains 4 classes, and report the performance in Tab. \ref{tab:50tasks_result}. SEMA outperforms all other methods in longer task sequences.
\begin{table}[ht]
	\centering
		\begin{tabular}{@{}lccccccccc cccccccc}
			\toprule
			\multicolumn{1}{l}{{Method}} & 
			\multicolumn{2}{c}{ImageNet-R}
        & \multicolumn{2}{c}{ImageNet-A} \\
		 & ${\bar{\mathcal{A}}}$ & {$\mathcal{A}_N $}
         & ${\bar{\mathcal{A}}}$ & {$\mathcal{A}_N $}
		\\
\midrule
L2P & 69.11          & 63.53          & 40.77          & 33.31       \\
DualPrompt                 & 64.21          & 56.25          & 49.74          & 39.83          \\
CODA-P                     & 61.34          & 56.37          & 34.36          & 23.17          \\
ADAM                       & 69.59          & 62.58          & 59.44          & 48.58          \\
InfLoRA                    & 67.01          & 61.37          & 47.33          & 31.27          \\
\midrule
SEMA                       & \textbf{74.64} & \textbf{67.03} & \textbf{60.82} & \textbf{49.18}       \\
\bottomrule
    \end{tabular}
          \caption{Evaluation on longer task sequence with 50 tasks.}
     \label{tab:50tasks_result}		
\end{table}

\subsection{Additional Results on Incremental Performance}
We present a comparison of performance across incremental stages for CIFAR-100, 20-Task ImageNet-R, 20-Task ImageNet-A and VTAB in Fig. \ref{fig:acc_record} of the main paper. We further conduct experiments on ImageNet-A which is split into 10 tasks. We provide the incremental performance of 5-Task ImageNet-R, 10-Task ImageNet-R and 10-Task ImageNet-A in Fig. \ref{fig:acc_record_supp}. Both figures show that SEMA performs consistently well with different dataset splits.

\subsection{Analyses on Training with Less Data}
\label{supp:less_data}
We further conduct analyses on the scenario of training with less data. Benefiting from the better knowledge reuse/transfer ability, SEMA can achieve better performance with less data. We specifically compare with a state-of-the-art method, EASE \cite{ease}, which expands task-specific adapters at all layers of the transformer. Unlike all other methods we compared with in the main paper, EASE also incrementally adds classification heads for all tasks and ensembles them in inference. In Tab. \ref{tab:data_efficient_vtab}, we show the results of experiments on VTAB while removing 90\% of samples in one and two tasks, respectively, denoted as VTAB-1 and VTAB-2. Although EASE uses a much stronger classification head, SEMA can perform better in this data efficiency learning experiment. We then further extend this data efficiency experiment to ImageNet-A by keeping only 10 or 20 percent of data for all tasks. As shown in Tab. \ref{tab:data_efficient_ina}, with sub-linear expansion, SEMA obtains performance comparable to EASE which requires task-oriented expansion at linear growth rate.

\begin{table}[htp]
\centering
\begin{tabular}{lcccccccc}
\toprule
\multirow{2}{*}{Method} & \multicolumn{2}{c}{VTAB-1}                                   & \multicolumn{2}{c}{VTAB-2}                                 \\
                        & {${\bar{\mathcal{A}}}$} & $\mathcal{A}_N $ & {${\bar{\mathcal{A}}}$} & $\mathcal{A}_N $ \\ \midrule
SEMA                    & 86.74                                     & 81.33            & 85.99                                     & 80.06                     \\
\midrule
EASE                    & 86.56                                     & 78.37            & 86.76                                     & 78.86         \\
\bottomrule
\end{tabular}
\caption{Experiments on setting with limited data samples on VTAB. VTAB-1 and VTAB-2 randomly removes 90 percent of data in one and two task(s), respectively.}
\label{tab:data_efficient_vtab}
\end{table}

\begin{table}[htp]
\centering
\begin{tabular}{lcccccccc}
\toprule
\multirow{2}{*}{Method}  & \multicolumn{2}{c}{ImageNet-A 10\%}                          & \multicolumn{2}{c}{ImageNet-A 20\%}                                           \\
                        & {${\bar{\mathcal{A}}}$} & $\mathcal{A}_N $ & {${\bar{\mathcal{A}}}$} & $\mathcal{A}_N $  \\ \midrule
SEMA                & 52.90                                     & 41.41            & 57.85                                     & 48.26            \\
\midrule
EASE                & 52.79                                     & 41.67            & 57.46                                     & 48.65      \\
\bottomrule
\end{tabular}
\caption{Experiments on setting with limited data samples on ImageNet-A. ImageNet-A 10\% contains only 10 percent of data in original ImageNet-A for all tasks and ImageNet-A 20\% contains 20 percent.}
\label{tab:data_efficient_ina}
\end{table}

\subsection{Experimental Results with Different Seeds and Varying Class Orders}
\label{supp:error_bar}
We conduct five independent runs with different seeds for SEMA on all datasets, and report the mean and standard deviation of accuracies over separate runs in Tab. \ref{tab:exp_std_result}. With different random seeds, each run is performed with different shuffling of class order and model initialization weights. This demonstrates the robustness of SEMA's performance with varying task/class orderings.

\begin{figure*}[!t]
  \centering
  \begin{subfigure}[b]{0.3\textwidth}
    \includegraphics[width=\linewidth]{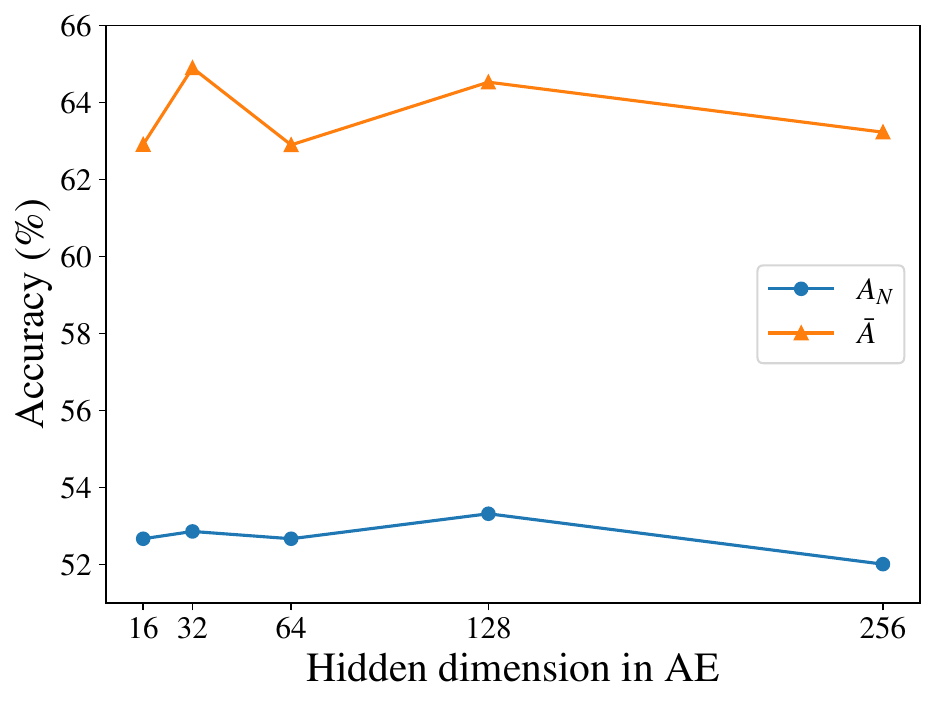}
    \caption{ImageNet-A}
    \label{fig:ina_ae_dim}
  \end{subfigure}
  \hspace{1cm}
  \begin{subfigure}[b]{0.3\textwidth}
    \includegraphics[width=\linewidth]{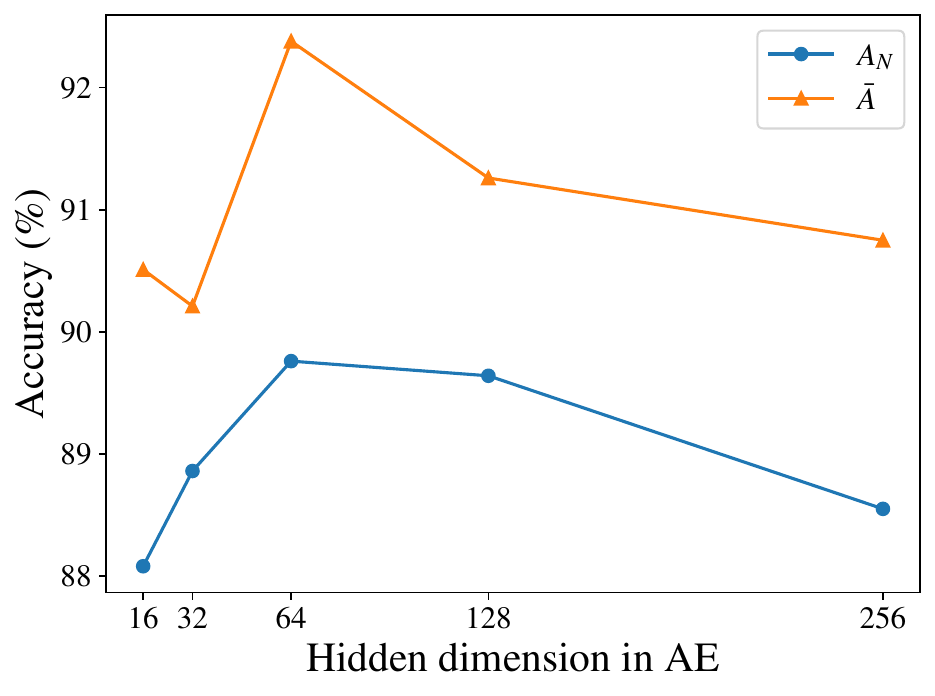}
    \caption{VTAB}
    \label{fig:vtab_ae_dim}
  \end{subfigure}
  \caption{Ablation on representation descriptor. 
  }
  \label{fig:ae_dim}
\end{figure*}

\begin{table*}[htp]
\centering
\begin{tabular}{llcccccc}
\toprule
\multicolumn{2}{l}{Method}                    & CIFAR-100        & 5-Task IN-R & 10-Task IN-R & 20-Task IN-R       & ImageNet-A       & VTAB             \\ \midrule
\multirow{2}{*}{SEMA} 
& ${\bar{\mathcal{A}}}$ & 91.37 $\pm$ 0.38 & 84.75 $\pm$ 0.84 & 83.56 $\pm$ 0.41                & 81.75 $\pm$ 1.00 & 64.53 $\pm$ 0.99 & 91.26 $\pm$ 0.47 \\
& $\mathcal{A}_N $      & 86.98 $\pm$ 0.57 & 79.78 $\pm$ 0.46 & 78.00 $\pm$ 0.49                   & 74.53 $\pm$ 0.92 & 53.32 $\pm$ 0.69 & 89.64 $\pm$ 0.63 \\
\bottomrule
\end{tabular}
\caption{Accuracies with standard deviation over 5 independent runs.}
\label{tab:exp_std_result}
\end{table*}

\subsection{Ablation Study on the Hidden Dimension in AE}
We test different values for hidden dimensions in the AE as representation descriptors. The AE-based representation descriptors enable the capture of the characteristics of the data for decision-making on whether to add a new adapter during continual training. According to Fig. \ref{fig:ae_dim}, the proposed method can perform well with a wide range of settings on the AE's hidden dimension.

\begin{table*}[!htp]
	\centering
        \begin{tabular}{@{}lccccccccc cccccccc}
        \toprule
        \multicolumn{1}{l}{{Method}} & 
        \multicolumn{2}{c}{CIFAR-100}
            & \multicolumn{2}{c}{5-Task IN-R}
            & \multicolumn{2}{c}{10-Task IN-R}
            & \multicolumn{2}{c}{ 20-Task IN-R} 
        & \multicolumn{2}{c}{ImageNet-A} & \multicolumn{2}{c}{VTAB} \\
		 & ${\bar{\mathcal{A}}}$ & {$\mathcal{A}_N $}
         & ${\bar{\mathcal{A}}}$ & {$\mathcal{A}_N $}
       & ${\bar{\mathcal{A}}}$  & {$\mathcal{A}_N $} 
        & ${\bar{\mathcal{A}}}$ & {$\mathcal{A}_N $}
       & ${\bar{\mathcal{A}}}$  & {$\mathcal{A}_N $} 
       & ${\bar{\mathcal{A}}}$   & {$\mathcal{A}_N $} 
    \\
    \midrule
RanPAC      & 93.81          & 90.04           & 83.81          & 79.57          & 84.23          & 79.00          & 83.87          & 78.18            & 69.96          & 62.15            & 91.97          & 91.33          \\
\midrule
SEMA+RanPAC & \textbf{94.54} & \textbf{90.95}  & \textbf{85.93} & \textbf{81.58} & \textbf{85.59} & \textbf{80.55} & \textbf{85.13} & \textbf{79.40}   & \textbf{71.87} & \textbf{63.33}   & \textbf{93.99} & \textbf{92.33} \\ 
\bottomrule
    \end{tabular}
    \caption{Results on different methods using random projection technique.}
     \label{tab:ranpac_results}	
\end{table*}

\subsection{Results with Representation Enhancement}
\label{supp:feat_align}
As discussed, different PTM-based continual learning methods focus on updating/adapting the backbone/representation (\eg, SEMA, InfLoRA~\cite{inflora}, CODA-P~\cite{coda}) and continually conducting feature representation enhancement of frozen PTMs (\eg, RanPAC \cite{ranpac}), respectively. These two types of methods are orthogonal and can work together. 
The proposed self-expansion learning in SEMA can also be combined with the statistical alignment techniques of RanPAC, \ie, SEMA+RanPAC, to get better performance. Specifically, the feature enhancement with random projection and prototype classifiers in RanPAC is applied to the representations from SEMA's model. Tab. \ref{tab:ranpac_results} demonstrates that the representations are benefited from the self-expansion strategy, as SEMA+RanPAC outperforms RanPAC implemented with a single adapter and first-session adaptation.

\begin{table}[h]
        \centering
        \begin{tabular}{@{}lccc ccc}
            \toprule
            \multicolumn{1}{l}{{\small Method}} & 
            \multicolumn{2}{c}{\small CIFAR-100} 
            & \multicolumn{2}{c}{\small 10-Task IN-R} \\
         & ${\bar{\mathcal{A}}}$ & {$\mathcal{A}_N $}
         & ${\bar{\mathcal{A}}}$ & {$\mathcal{A}_N $}
        \\
        \midrule
    Zero-shot & 76.36  & 66.96  & 79.17 & 77.08 \\
    ADAM                     & 79.53  & 71.26  & 72.06 &     70.90       \\
    \midrule
    SEMA        &  \textbf{82.74}    & \textbf{73.52}     &  \textbf{80.94}       & \textbf{78.18}       \\
    \bottomrule
        \end{tabular}
        \caption{Performance on pre-trained CLIP model.}
         \label{tab:clip_result}	
\end{table}
\subsection{Experiments with CLIP.}
We further conduct experiment with a pre-trained vision-language model, namely CLIP with a ViT-B/16 backbone~\cite{clip}, and report the performance in Tab. \ref{tab:clip_result}. SEMA outperforms zero-shot CLIP and ADAM which have no parameter expansion, highlighting the effectiveness of our dynamic expansion strategy and its broad applicability to pre-trained models.

\end{document}